%% file: arxive.tex
\documentclass{article} 
\usepackage{iclr2027_conference,times}

\input{math_commands.tex}

\usepackage{comment}
\usepackage{url}
\usepackage{amsmath}
\usepackage{amssymb}
\usepackage{float}
\usepackage{booktabs}
\usepackage{graphicx}
\usepackage[table]{xcolor}
\usepackage{tikz}
\usepackage{subcaption}
\usepackage[most]{tcolorbox}
\usepackage{hyperref}

\usetikzlibrary{arrows.meta,positioning}

\usepackage{needspace}
\let\oldsection\section
\renewcommand{\section}{\needspace{5\baselineskip}\oldsection}

\newcommand{\Gstruct}{\mathcal{G}_{S}}
\newcommand{\Glat}{\mathcal{G}_{L}}
\def\our{STAGE}

\title{\our{}: Subspace-Targeted Affine Generative Erasure for Text-to-3D Models}

\iclrfinalcopy

\author{Karol Dziekan$^{1}$, Przemys{\l}aw Spurek$^{1,2}$,  Dawid Malarz$^{1,2}$\\
Jagiellonian University$^1$; 
IDEAS Research Institute$^2$ \\
\texttt{\{dziekan,spurek,malarz\}@uj.edu.pl} \\
} 

\begin{document}

\maketitle

\begin{abstract}
Concept erasure suppresses a target concept while preserving behavior on unrelated inputs. Existing closed-form methods were designed for 2D image diffusion and assume a single generative pathway, so one edit must cover geometry and texture at once. Native 3D generators, which synthesize structured 3D representations directly rather than by lifting 2D samples, violate this assumption. We show that shape and object concepts must be erased in the structural stage of the pipeline and material concepts in the appearance stage. We therefore formulate erasure in native text-to-3D as a stage-aware editing problem and introduce \our{}, a training-free, closed-form framework. \our{} confines each edit to the low-dimensional subspace spanned by the differences between erase and anchor embeddings, and relaxes the norm-preserving (orthogonal) constraint of prior editors into a least-squares affine correction that maps target activations onto safe anchors subject to a penalty on the displacement of retained prompts. The correction applies to the structural stage, the appearance stage, or both. We find that the stage an edit must reach is determined by concept type. On TRELLIS, the standard open native 3D generator, across 15 shape, material, and object concepts, \our{} reaches 66.7 on a composite score that balances forgetting the target concept against preserving everything else, aggregating CLIP-based semantic and physical metrics, versus 53.2 for the strongest adapted baseline.\\
    Code: \url{https://github.com/gmum/STAGE/} \\
    Project Page \url{https://gmum.github.io/STAGE/}
\end{abstract}

\begin{figure}[h]
  \centering
  \includegraphics[width=\textwidth]{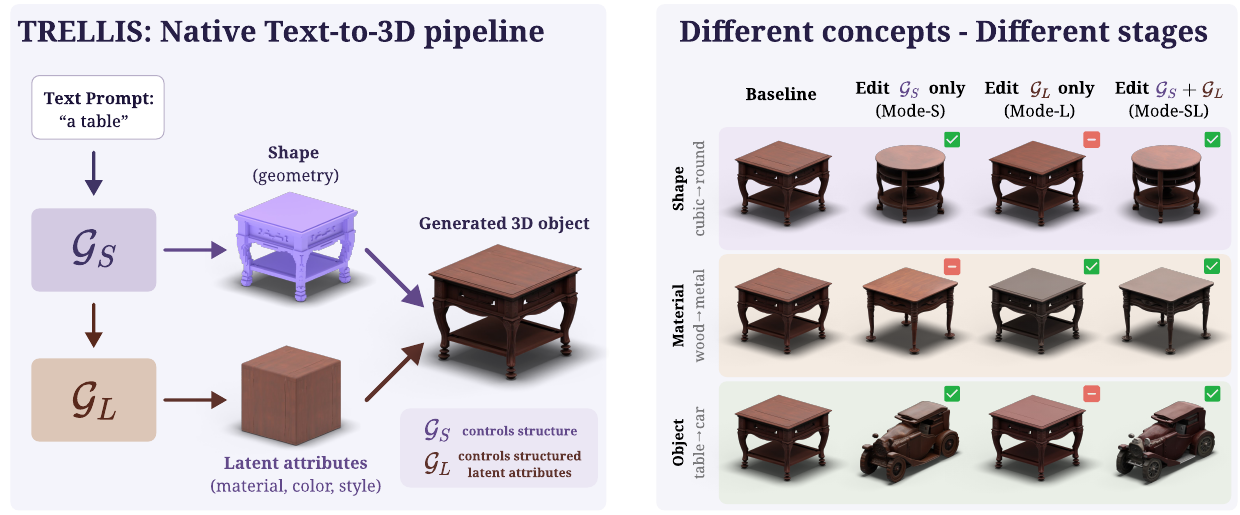}
  \caption{\textbf{Concepts localize to different stages of TRELLIS's pipeline.} Left: TRELLIS generates an asset in two stages, a structural stage $\Gstruct$ that fixes the voxel scaffold and a latent stage $\Glat$ that fills it with material, color, and fine geometry. Right: editing only $\Gstruct$ (Mode-S) moves shape but not material. Editing only $\Glat$ (Mode-L) moves material but not shape. Object identity already shifts under Mode-S alone, but the full appearance change needs editing both stages, which is why \our{} assigns object concepts to Mode-SL.}
  \label{fig:teaser}
\end{figure}

\section{Introduction}

Native 3D generators synthesize structured 3D representations directly rather than by lifting 2D samples, and now produce high-fidelity assets \citep{xiang2025trellis,zhao2025hunyuan3d,zhang2024clay}, building on advances in latent diffusion and rectified flows \citep{rombach2022ldm,liu2023rectified}. As these models enter content-creation pipelines, they must be prevented from generating copyrighted, sensitive, or harmful content. In 2D image diffusion, this is the goal of concept erasure, which suppresses a target concept while preserving behavior on unrelated inputs \citep{gandikota2023esd, kumari2023ablating}. Prior 3D unlearning work \citep{wang2025unlearning3d} targets image-to-3D novel-view synthesis models, which output multi-view images rather than 3D representations, and fine-tunes them to forget individual input objects. Erasing text-specified concepts from native text-to-3D generators remains largely unaddressed.

Training-free closed-form editors avoid the cost of fine-tuning large 3D generators, but were designed for 2D image diffusion, where a single generative pathway means one edit must cover geometry and texture at once (Section~\ref{sec:background}). Native 3D generators violate this assumption. TRELLIS~\citep{xiang2025trellis} first generates a sparse voxel structure in a structural stage and then, in an appearance stage, generates structured latents on the occupied voxels that encode appearance and fine local geometry, each read through its own cross-attention layers. We show that shape, material, and object concepts depend on different stages of this pipeline. Editing only the appearance stage fails for shape and object concepts, whereas material concepts are erased more effectively there than in the structural stage, and most completely when both stages are edited (Section~\ref{sec:mode-assignment}). This raises the cost of imprecise edits. An edit applied to a stage that does not carry the target concept can only cause collateral change, and collateral changes in the structural stage propagate to the appearance stage through the scaffold it produces.

We therefore formulate concept erasure in native text-to-3D generation as a stage-aware editing problem and introduce \our{} (Figure~\ref{fig:teaser}), a training-free, closed-form framework for it. \our{} confines each edit to the low-dimensional subspace spanned by the differences between erase and anchor embeddings, then fits a least-squares affine correction that reads its input only through that subspace and maps target activations onto safe anchors, with a penalty on the displacement of retained prompts (Section~\ref{sec:mods}). Because both stages read the same text embeddings, the same correction can be folded exactly into the cross-attention layers of the structural stage, the appearance stage, or both, and we place it according to concept type.

Our contributions are as follows:
\begin{itemize}
    \item A stage-aware formulation of concept erasure for native text-to-3D generation, together with an analysis on TRELLIS showing that the stage an erasure must reach depends on concept type.
    \item \our{}, a closed-form editor that combines subspace confinement with a least-squares affine correction, folded exactly into the cross-attention layers of either stage or both.
    \item An evaluation on TRELLIS across 15 shape, material, and object concepts. With the same stage placement for all editors, \our{} reaches 66.7 on the 3D Unlearning Score, a composite score we introduce that aggregates CLIP-based semantic and physical metrics, compared with 53.2 for the strongest adapted baseline. Even at its default hyperparameters, \our{} scores above both tuned baselines (60.1 vs 53.2).
\end{itemize}

\section{Related Work}
\label{sec:background}

Concept-erasure methods differ mainly in when and how they intervene in a pretrained generator. Inference-time methods leave weights untouched, either by steering the sampling trajectory, as in Safe Latent Diffusion~\citep{schramowski2023sld}, or by projecting cross-attention values away from the erased concept, as in AdaVD~\citep{wang2025adavd}. Training-based methods fine-tune selected parameters against an erasure objective, including ESD~\citep{gandikota2023esd} and later fine-tuning approaches~\citep{kumari2023ablating,lu2024mace}. Closest to our setting, model-editing methods update a small set of text-conditioning layers in closed form, without gradient-based fine-tuning. UCE~\citep{gandikota2024uce} solves a least-squares problem that redirects erased concepts while preserving a retain set. RECE~\citep{gong2024rece} applies such updates repeatedly to newly derived adversarial embeddings. OCE~\citep{sun2026oce} replaces the least-squares update with a norm-preserving rotation obtained from the orthogonal Procrustes problem~\citep{schonemann1966procrustes}. As we show in Section~\ref{sec:mods}, a rotation preserves distances and therefore cannot map an erased concept onto a distinct anchor without also moving that anchor. \our{} relaxes this constraint to a least-squares affine correction confined to the subspace spanned by the differences between erase and anchor embeddings. This differs from UCE, whose least-squares update is unconstrained.

The methods above were all designed for 2D image diffusion. The only prior 3D-unlearning method we are aware of~\citep{wang2025unlearning3d} fine-tunes the 2D multi-view diffusion model at the front of an image-to-3D pipeline and leaves the downstream reconstruction stage untouched. \our{} instead edits the text-conditioning layers of a native 3D generator directly, including those of the structural stage, which fixes coarse geometry before appearance is generated and has no counterpart in a 2D pipeline.

Evaluating concept erasure also changes once the output is a 3D asset rather than an image. Prior work scores 2D erasure with classifier accuracy, CLIP-based similarity~\citep{hessel2021clipscore}, or distributional metrics such as FID~\citep{heusel2017fid}, often summarized with a harmonic mean~\citep{sun2026oce}. Applied to rendered views alone, such scores depend on the chosen cameras and cannot tell whether a change comes from geometry or from appearance. Section~\ref{sec:3dus} introduces the 3D Unlearning Score, which pairs CLIP scores on rendered views with a point-cloud evaluator for shape and object concepts and a material classifier for material concepts, and measures preservation relative to the unedited model.

\section{Method}
\label{sec:mods}

\begin{figure}[H]
  \centering
  \includegraphics[width=\textwidth]{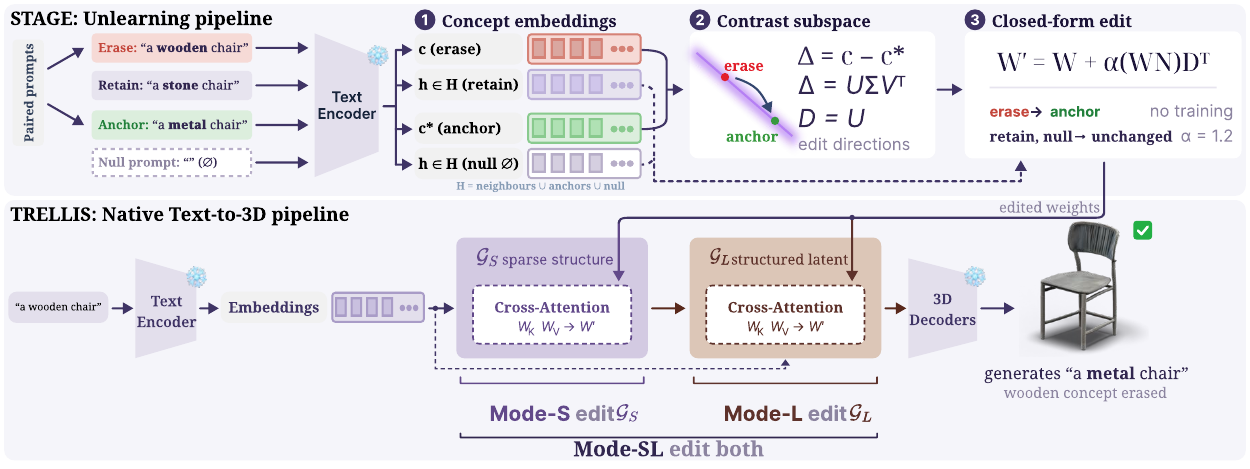}
  \caption{%
    \textbf{\our{} folded into TRELLIS.} Top: \our{} encodes the paired prompts, including the retained set $\mathcal{H}$ of anchors, the null prompt, and optional neighbors, builds the contrast subspace $D$ from the differences between erase and anchor embeddings, and solves the closed-form correction in one step. Bottom: the resulting weights replace $W_K$ and $W_V$ in the cross-attention of $\Gstruct$, $\Glat$, or both, depending on the concept, with no retraining.%
  }
  \label{fig:pipeline}
\end{figure}

\paragraph{Preliminaries and Problem Formulation}
\label{sec:background-trellis}
Following TRELLIS~\citep{xiang2025trellis}, we write the generator as two cascaded stages, a structural stage $\Gstruct$ and a latent appearance stage $\Glat$. A text prompt is encoded by a frozen text encoder~\citep{radford2021clip} into an embedding $c$. 
 Given $c$, the structural stage first constructs the underlying geometric scaffold (the active voxel support), and the appearance stage then generates appearance and fine local geometry on that fixed support. Both stages read the same embedding independently, each through its own cross-attention layers. Because $\Gstruct$ and $\Glat$ are separate networks, an edit can be folded into either one, or both, without changing the other's weights. Figure~\ref{fig:pipeline} shows the full procedure.

Given a pretrained generator and a set of concepts to erase $C_\text{erase}=\{c_1,\dots,c_k\}$, we assign each one a safe anchor concept, $C_\text{anchor}=\{c^*_1,\dots,c^*_k\}$. The goal of training-free weight editing is to find updated weights $W^*$ so that the model maps any prompt built around $c_i$ to the same representation it would already produce for $c^*_i$, while leaving its behavior on every unrelated concept unchanged.

\paragraph{Stage Placement}
\label{sec:method-modes}
\our{} exploits this split between geometry and appearance by choosing, per concept, which stage's cross-attention layers receive the edit. Inside each block, the key and value projections are a single fused linear layer. \our{} always edits both halves together, so a redirected concept is read consistently whether it is used as a key or as a value. This gives three modes:
\begin{itemize}
    \item \textbf{Mode-S:} fold the correction only into $\Gstruct$'s layers, leaving appearance untouched (assigned to shape concepts).
    \item \textbf{Mode-L:} fold the correction only into $\Glat$'s layers, leaving structure untouched (assigned to material concepts).
    \item \textbf{Mode-SL:} fold the correction into both (assigned to object concepts, whose identity depends on both where matter is and what it looks like).
\end{itemize}
Section~\ref{sec:mode-assignment} assigns each concept axis its mode from this reasoning and tests the assignment empirically. In the main comparison (Section~\ref{sec:experiments}), all editors, including \our{}, use this per-axis assignment, so that all methods edit the same layers and the reported differences reflect the edit rather than its placement.

\paragraph{Subspace Localization}
The updates of UCE and OCE are not restricted to a subspace, so any input can be affected. UCE constrains the change only on the concepts in its preservation term, and OCE's rotation acts on the whole output space of the projection. In a cascaded 3D architecture this matters more than in a single network. An unnecessary change in how $\Gstruct$ reads an unrelated prompt alters the voxel support on which $\Glat$ then generates appearance, so a small perturbation early in the cascade becomes a visible change in the final asset.

\our{} instead restricts the correction to a small, data-driven subspace built directly from the erase and anchor pairs, rather than the whole embedding space. We write $C_\text{erase}$ and $C_\text{anchor}$ to be \emph{template-parallel}. Erase prompt $i$ and its anchor prompt differ only in the concept word itself, so the difference between their text embeddings is dominated by the directions that encode that concept. We pair their text-encoder outputs at every aligned token position rather than only their last-token summary, so each prompt pair contributes several per-token difference vectors. From here on, $c_i$ and $c^*_i$ denote the $i$-th aligned pair of erase and anchor token states, and $k$ the number of such pairs. Collecting the differences $\delta_i=c_i-c^*_i$ into a matrix $\Delta=[\delta_1\,\cdots\,\delta_k]$ and taking its SVD, the leading $r$ left singular vectors $D\in\mathbb{R}^{d\times r}$ span a subspace $V$ along which erase and anchor concepts actually differ, $r{=}16$ dimensions by default
out of the full embedding width $d{=}768$. The correction we fit below depends on an input only through its projection $D^\top c$ onto $V$, plus one constant offset shared by every input. An unrelated prompt is therefore displaced only through its projection onto $V$ and the shared offset, and the objective below penalizes this displacement on a retained set of prompts we want left alone. Because the offset is shared by every input, this penalty keeps it small wherever it would move retained prompts.

\begin{figure}[h]
\centering
\includegraphics[width=\textwidth]{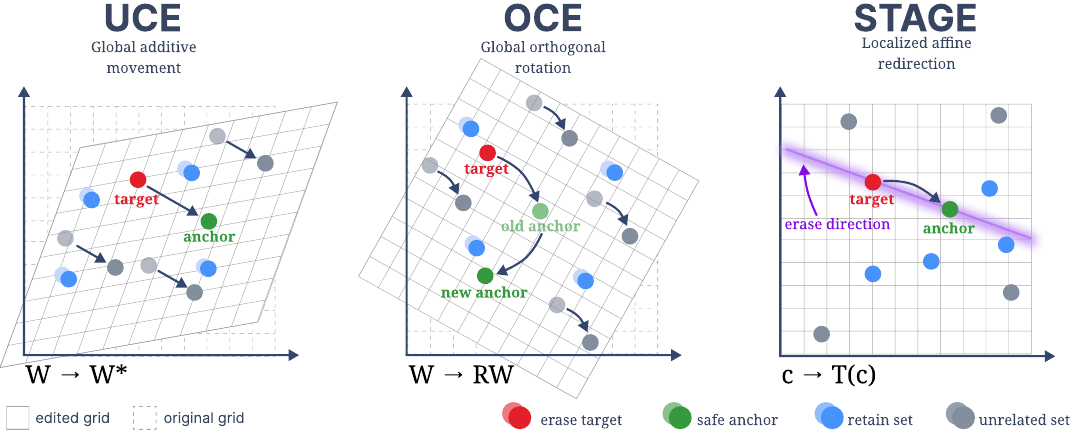}
\caption{\textbf{Geometric intuition of the editing primitives.} Left: a least-squares update to the full projection (UCE-style) is not confined to a subspace, so prompts outside the preserved set can drift. Middle: a rotation $R$ (OCE-style) preserves distances, so moving $c$ onto $c^*$ forces $c^*$ itself to move to $Rc^*$. Right: \our{}'s correction reads an input only through its projection $D^\top c$ onto $V=\mathrm{span}(D)$, plus one offset shared by every input. The retained-set penalty of equation~(\ref{eq:stage_obj}) keeps the displacement of anchors and retained prompts small, so $T(c^*)\approx c^*$.}
\label{fig:geometric_intuition}
\end{figure}

\paragraph{Closed-Form Affine Correction}
Rotation-based methods such as OCE~\citep{sun2026oce} search for a rotation matrix aligning erase and anchor embeddings. But a rotation is a distance-preserving isometry. It keeps the distance between a target and its anchor fixed, so it cannot map the target onto the anchor while leaving the anchor in place. \our{} instead fits an \emph{affine} correction that acts through the subspace $V$ above, $T(c) = c + N(D^\top c) + n_0$, where $N\in\mathbb{R}^{d\times r}$ maps the subspace projection back into the full embedding space and $n_0\in\mathbb{R}^d$ is a constant offset. An affine map need not preserve distances, so it can merge two distinct points. Figure~\ref{fig:geometric_intuition} contrasts the three editing primitives.

We fit $N$ and $n_0$ by regularized least squares. $T$ should map every erase embedding onto its anchor, while a retained set $\mathcal{H}$, namely the anchor embeddings themselves, the model's null (unconditional) prompt, and optionally a handful of concept-specific neighbors, should move as little as possible, each weighted by $\beta$ (the null prompt more heavily, since classifier-free guidance amplifies any drift there), and a ridge term $\lambda$ keeps the fit well-conditioned:
\begin{equation}
\min_{N,n_0}\ \sum_{i=1}^k \big\lVert N(D^\top c_i)+n_0 - (c^*_i-c_i)\big\rVert_2^2
\;+\;\beta\!\!\sum_{h\in\mathcal H}\!\big\lVert N(D^\top h)+n_0\big\rVert_2^2
\;+\;\lambda\big(\lVert N\rVert_F^2+\lVert n_0\rVert_2^2\big).
\label{eq:stage_obj}
\end{equation}
With thousands of token pairs per concept, $\lambda$ mainly guards against ill-conditioning and has little effect on the fit (Table~\ref{tab:ablation-stage-full}). A further term, $\gamma\,\mathrm{tr}\big(N(D^\top K_0 D)N^\top\big)$, regularizes $N$ against the corpus-level second moment $K_0=\mathbb{E}_c[cc^\top]$ of TRELLIS's own training captions, restricted to $V$ by $D$. It discourages corrections that align with directions of unusually high variance in the prompt distribution. We estimate $K_0$ once on TRELLIS's short training-style captions rather than on natural photo captions, since a corpus mismatch would bias it systematically rather than just add noise. We omit this term from equation (\ref{eq:stage_obj}) for readability and ablate $\gamma$ in Section~\ref{sec:ablations}. Equation (\ref{eq:stage_obj}) is a strictly convex quadratic in $(N,n_0)$, so it has a unique closed-form minimizer, obtained from one small $(r{+}1)\times(r{+}1)$ linear solve, independent of the embedding width $d$, and reused unchanged for every layer the edit is folded into, since $D$, $N$, and $n_0$ depend only on text embeddings, never on any layer's own weights.

Folding $T$ into a cross-attention projection $(W,b)$ is exact rather than approximate. Since $W\,T(c)+b=Wc+(WN)(D^\top c)+Wn_0+b$ for every $c$, setting
\begin{equation}
W_{\text{new}} = W + (WN)D^\top, \qquad b_{\text{new}} = b + Wn_0
\label{eq:stage_fold}
\end{equation}
reproduces $W\,T(c)+b$ exactly, with no retraining. The entire edit is a handful of small matrix products, which is why fitting and applying it to every targeted layer of a TRELLIS stage takes seconds rather than a training run. The deployed correction is scaled by an overshoot factor $\alpha$. We fold $\alpha N$ and $\alpha n_0$ in place of $N$ and $n_0$, so that $T_\alpha(c)=c+\alpha\,(N D^\top c+n_0)$, which for $\alpha>1$ pushes erased concepts slightly past their anchors. We study its effect in Section~\ref{sec:ablations}.

\section{Experiments}
\label{sec:experiments}

\paragraph{Experimental Setup}
\label{sec:setup}

We evaluate \our{} on the native 3D generator \texttt{microsoft/TRELLIS-text-xlarge}, using 50 rectified-flow sampling steps per stage and keeping the CLIP text encoder and the structured-latent (SLat) decoders frozen. We curate 15 concepts, five on each of three axes: \emph{shape} (sphere, cube, cylinder, cellular lattice, torus), \emph{material} (wood, metal, glass, stone, ceramic), and \emph{object identity} (car, cat, chair, table, teddy bear).

Each concept uses three mutually disjoint prompt sets. The edit is fitted on paired, template-parallel erase and anchor prompts that differ only in the concept word, with the retained set $\mathcal{H}$ of equation~(\ref{eq:stage_obj}) built from the anchor embeddings and the null prompt. Evaluation uses a separate set of 30 unseen prompts per concept. Erasure can only be assessed if the base model produces the concept in the first place, so each evaluation prompt is fixed to a seed for which every evaluator of its axis recognizes the concept on the unedited model (Appendix~\ref{sec:app-data}). All methods and ablations reuse these (prompt, seed) pairs, which removes sampling noise from comparisons between them.

Edit modes follow the per-axis assignment of Section~\ref{sec:method-modes}. Every method uses its best configuration from the same one-at-a-time hyperparameter sweep, given in Table~\ref{tab:ablation-stage} for \our{} and in Appendix~\ref{sec:baseline-tuning} for UCE and OCE. The baselines, UCE and OCE, use the same prompts, seeds, and per-axis stage assignment (Appendix~\ref{sec:app-impl}), and each edits the projections that give it its highest score, which for UCE are the key and value projections of its original implementation (Table~\ref{tab:ablation-kv}).

\paragraph{The 3D Unlearning Score}
\label{sec:3dus}

A score for 3D erasure has to answer two questions at once: whether the target
concept left, and whether everything else survived. We measure both per concept,
with evaluators chosen for the axis that concept belongs to, and combine them so
that a gain on one cannot hide a loss on the other.

\emph{Forgetting.}
Every concept is scored by two evaluators, one semantic and one physical (Appendix~\ref{sec:app-metrics}). CLIP
\citep{radford2021clip} is the semantic evaluator throughout. Shape and object
concepts are additionally scored on point clouds with Uni3D~\citep{zhou2024uni3d},
and material concepts with MINC-23~\citep{bell2015minc}, restricted and
renormalized over our five material classes. All evaluators choose among the same closed candidate set $\mathcal{C}$, the concepts on that concept's own axis (five in our evaluation), so they share a chance level $\rho_e = 100/|\mathcal{C}|$,
and the detection rate $E_{e,c}$, the percentage of the evaluation assets of $c$ in which evaluator $e$ detects $c$, becomes a chance-adjusted rate
$\hat{E}_{e,c} = \mathrm{clip}\big((E_{e,c}-\rho_e)/(100-\rho_e),\,0,\,1\big)$. For the
image-based evaluators, an asset counts as detected if the concept wins in at least one of
$N_v=4$ views spread on a Fibonacci sphere, so a concept counts as present if any
view reveals it. We define forgetting as what stays absent after the edit,
$F_{e,c} = 1-\hat{E}^{\mathrm{post}}_{e,c}$. A concept is included in the evaluation only if the unedited model expresses it reliably, $\hat{E}^{\mathrm{pre}}_{e,c}\ge\tau=0.5$ for every evaluator of its axis, which holds for all 15 concepts.

\emph{Preservation.}
Classification accuracy is too coarse for the other half. It does not
move until a prediction flips, and an edit can visibly damage an asset long before
that. We score preservation continuously instead. For a non-target concept
$c'\in\mathcal{C}^{-}_c$, one of the other concepts on the axis of $c$, and an asset $x$ generated for a prompt of $c'$, let $A_e(x,c')$ be the cosine similarity between the evaluator's embedding of $x$ and the text of that prompt, and $\bar{A}_e(x,c')$ the mean similarity of $x$ to the labels of the other concepts on the axis. We center the first on the second,
$\hat{A}_e(x,c') = \mathrm{clip}\big((A_e(x,c')-\bar{A}_e(x,c'))/(1-\bar{A}_e(x,c')),\,0,\,1\big)$,
average it over the evaluation assets of $c'$ to obtain $\hat{A}_{e,c'}$,
and compare this mean after the edit against the unedited model, with a small $\epsilon$ for numerical stability:
\begin{equation}
P_{e,c} = \min\Big(1,\; \frac{1}{|\mathcal{C}_c^{-}|}\sum_{c'\in\mathcal{C}_c^{-}}
\hat{A}^{\mathrm{post}}_{e,c'}\big/\big(\hat{A}^{\mathrm{pre}}_{e,c'} + \epsilon\big)\Big).
\label{eq:3dus_preservation}
\end{equation}
For MINC-23, $\hat{A}$ is the chance-calibrated classifier confidence in the asset's own material
rather than a cosine similarity. Preservation is therefore measured against the
unedited backbone rather than in absolute terms, and capped at 1 so that an edit
earns no credit for improving unrelated concepts.

\emph{Combining the two.}
For each concept, the two evaluators of its axis, written $\mathcal{E}_a$ (CLIP plus Uni3D or MINC-23), are combined by geometric mean,
$F_c = \big(\prod_{e\in\mathcal{E}_a} F_{e,c}\big)^{1/|\mathcal{E}_a|}$ and likewise
for $P_c$, which denies credit to an edit that removes a concept under one modality
while leaving it detectable under the other. We report the mean of the concept-level score over the five concepts of each axis and summarize the three axes by their geometric mean. The concept-level score itself is the harmonic mean of $F_c$ and $P_c$, so neither can be traded away for the other:
\begin{equation}
\mathrm{3D\mbox{-}US}_c = 100 \cdot \frac{2F_c P_c}{F_c + P_c + \epsilon}.
\label{eq:3dus}
\end{equation}

\begin{figure}[H]
  \centering
  \includegraphics[width=\linewidth]{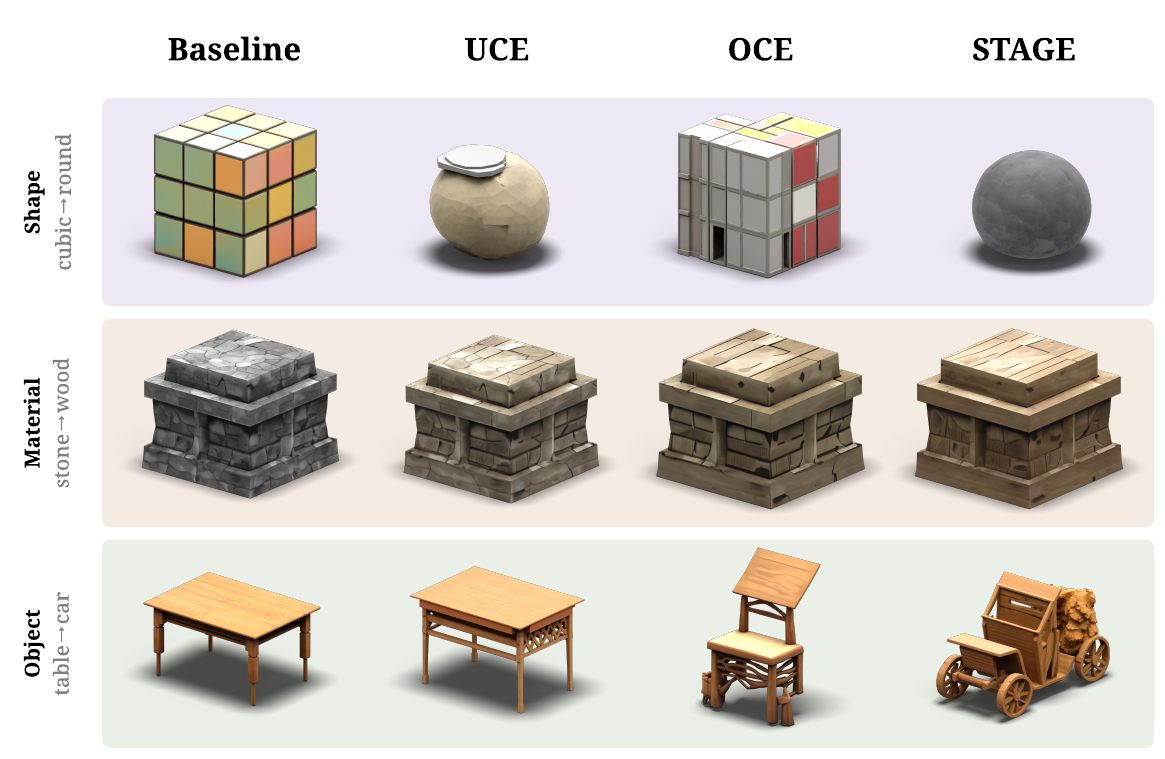}
  \caption{\textbf{Behavior of each editing method on one concept per axis.}
    Rows: shape (\texttt{cubic}$\to$\texttt{round}), material (\texttt{stone}$\to$\texttt{wood}), and object identity (\texttt{table}$\to$\texttt{car}). Columns: the unedited backbone, followed by UCE, OCE, and \our{}, all on the same prompt and seed. Each row shows
    how far the target concept moves toward its anchor and how much of the
    surrounding asset changes with it.}
  \label{fig:grid}
\end{figure}

\paragraph{Main Results}
\label{sec:main-results}

\begin{table}[H]
\centering
\small
\setlength{\tabcolsep}{5pt}
\caption{
3D unlearning performance on TRELLIS. Each method is reported at its best
configuration under the tuning protocol of Appendix~\ref{sec:baseline-tuning}, and all
methods use the same per-axis stage assignment. $F$ measures target forgetting,
$P$ measures non-target preservation, and 3D-US combines both objectives. Axis
values are means over the axis's five concepts, and Overall is the geometric mean of
the three axis scores. $F$ and $P$ lie in $[0,1]$ and 3D-US in $[0,100]$. Higher values are better, and bold marks the best value in each column.
}
\label{tab:3dus}
\begin{tabular}{l ccc ccc ccc c}
\toprule
&
\multicolumn{3}{c}{Shape}
&
\multicolumn{3}{c}{Material}
&
\multicolumn{3}{c}{Object}
&
Overall
\\
\cmidrule(lr){2-4}
\cmidrule(lr){5-7}
\cmidrule(lr){8-10}
\cmidrule(lr){11-11}
Method
& $F$ & $P$ & 3D-US
& $F$ & $P$ & 3D-US
& $F$ & $P$ & 3D-US
& 3D-US
\\
\midrule
UCE
& 0.61 & 0.85 & 66.9 & 0.22 & 0.84 & 30.8 & 0.57 & \textbf{0.65} & 51.9 & 47.5\\
OCE
& \textbf{0.73} & 0.67 & 67.9 & 0.36 & 0.74 & 41.8 & 0.83 & 0.41 & 53.2 & 53.2\\
\our{}
& 0.70 & \textbf{0.91} & \textbf{77.7} & \textbf{0.45} & \textbf{0.92} & \textbf{56.4} & \textbf{0.90} & 0.61 & \textbf{67.7} & \textbf{66.7}\\
\bottomrule
\end{tabular}
\end{table}

\begin{figure}[H]
  \centering
  \includegraphics[width=0.97\linewidth]{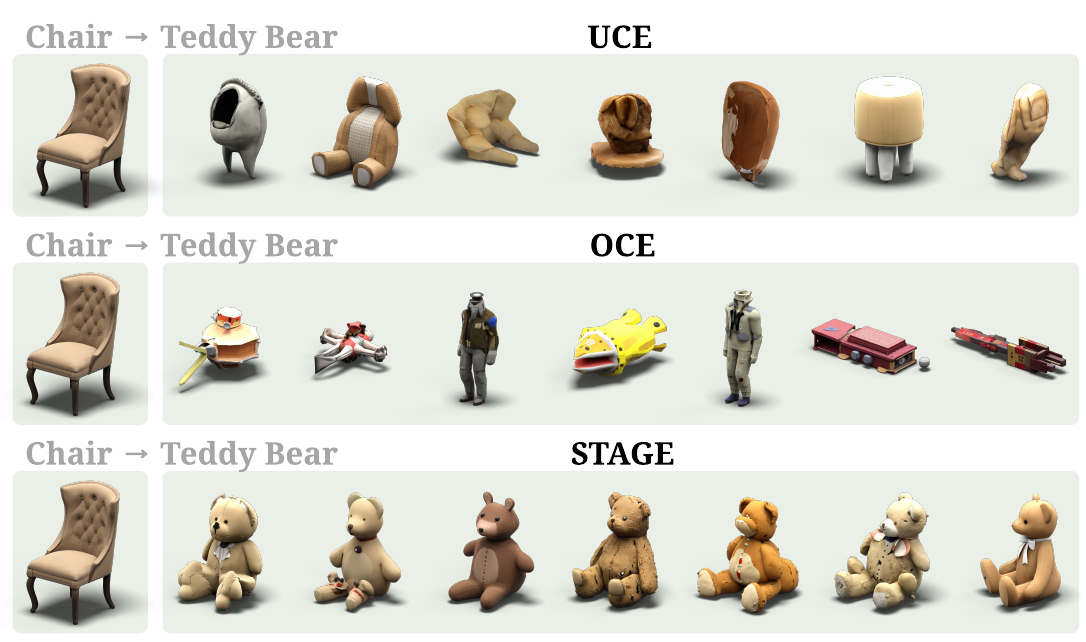}
  \caption{\textbf{Redirection across random seeds.} Erasing \texttt{chair} toward \texttt{teddy bear} on one evaluation prompt, for the unedited model (left) and seven seeds per editor. \our{} yields a teddy bear under every seed, while UCE and OCE mostly produce distorted or unrelated objects.}
  \label{fig:seeds}
\end{figure}

Table~\ref{tab:3dus} reports 3D-US, together with its forgetting ($F$) and preservation ($P$) components, for each method at its best configuration from the hyperparameter sweeps described below, with the per-axis mode assignment of Section~\ref{sec:method-modes}. \our{} attains the highest 3D-US on every axis and overall, with a margin of at least 9.8 points over OCE, the strongest baseline, on every axis. \our{} has the highest 3D-US on 10 of the 15 concepts, and on 9 of them it matches or exceeds both baselines in forgetting and in preservation at the same time, so its advantage does not come from trading one objective for the other. \our{} keeps preservation above 0.84 on every shape and material concept, whereas OCE drops as low as 0.57. The object axis is the hardest to preserve for every method. OCE falls below 0.5 on all five objects and \our{} on two of them, cat and teddy bear, while staying above 0.82 on the other three. UCE preserves objects best (0.65) but forgets far less than \our{} (0.57 vs 0.90). Appendix~\ref{sec:app-perconcept} lists the values for every concept. \our{} also redirects erased concepts to their anchors most often. Averaged over concepts and evaluators, 54.9\% of the edited evaluation assets are recognized as the anchor, compared with 33.7\% for UCE and 30.0\% for OCE, and at least 80\% on the object axis (Appendix~\ref{sec:app-perconcept}). Figures~\ref{fig:grid} and~\ref{fig:seeds} show qualitative examples, and Appendix~\ref{sec:app-qualitative} shows more.

\paragraph{Mode Assignment and Stage Locality}
\label{sec:mode-assignment}

\begin{table}[h]
\centering
\small
\caption{Mode crossover for all three methods at their tuned hyperparameters.
3D-US per axis under each of the three edit modes. Bold marks the mode assigned
to each axis in Table~\ref{tab:3dus}, and those cells are the values reported
there.}
\label{tab:mode-crossover}
\begin{tabular}{llccc}
\toprule
Method & Axis & Mode-S ($\Gstruct$ only) & Mode-L ($\Glat$ only) & Mode-SL (both) \\
\midrule
\our{} & Shape    & \textbf{77.7} & 0.0 & 77.2 \\
       & Material & 31.4 & \textbf{56.4} & 72.9 \\
       & Object   & 66.7 & 2.7 & \textbf{67.7} \\
\midrule
OCE    & Shape    & \textbf{67.9} & 3.0 & 56.5 \\
       & Material & 27.5 & \textbf{41.8} & 40.7 \\
       & Object   & 62.4 & 2.7 & \textbf{53.2} \\
\midrule
UCE    & Shape    & \textbf{66.9} & 0.0 & 56.3 \\
       & Material & 24.7 & \textbf{30.8} & 48.4 \\
       & Object   & 45.6 & 0.0 & \textbf{51.9} \\
\bottomrule
\end{tabular}
\end{table}

Rather than search over edit modes for each concept, we fix the mode per axis from the architectural role of each stage (Section~\ref{sec:method-modes}). Table~\ref{tab:mode-crossover} tests this assignment by scoring every axis under all three modes, with each method at its tuned hyperparameters. The structural stage is required for shape and object concepts. Editing only $\Glat$ collapses 3D-US on both axes (0.0 and 2.7 for \our{}), while editing only $\Gstruct$ already reaches 77.7 and 66.7. Adding $\Glat$ changes little for shapes (77.2) and completes the object edit (67.7) by also changing the appearance of the object (Figure~\ref{fig:teaser}). Material concepts are carried mainly by the appearance stage. Editing only $\Glat$ erases them far better than editing only $\Gstruct$ (56.4 vs 31.4), and editing both stages goes further (72.9), which suggests that materials also leave a trace in the voxel structure. We keep the assignment derived from stage roles, which edits the stage each concept type requires, rather than the highest-scoring mode. The advantage of \our{} does not depend on this choice. It scores highest of the three editors under every mode that erases a concept at all, and editing both stages on every axis gives an overall 3D-US of 72.5 for \our{}, compared with 49.6 for OCE and 52.1 for UCE.

\paragraph{Ablations}
\label{sec:ablations}

We vary the hyperparameters of \our{} one at a time around the default configuration, on all 15 concepts. Table~\ref{tab:ablation-stage} reports the best value of each hyperparameter, and Table~\ref{tab:ablation-stage-full} in Appendix~\ref{sec:baseline-tuning} gives the full sweep. Removing the subspace or the offset $n_0$, or using a rotation in the subspace, lowers overall 3D-US at the default configuration from 60.1 to between 41.3 and 54.1, while fitting only on the last token of each prompt, as UCE and OCE do, still gives 59.0 (Table~\ref{tab:ablation-components}).

Among $r$, $\beta$, the null-prompt multiplier, $\gamma$, and $\lambda$, the best single change ($r{=}32$) improves overall 3D-US by only 3.0 points over the default, and across the full sweep overall 3D-US stays between 57.5 and 63.1, although $\beta$ shifts the balance between the shape and object axes. Every configuration in this range already scores above the best tuned baseline (53.2 for OCE in Table~\ref{tab:3dus}). The overshoot scale $\alpha$ is the only hyperparameter with a large effect. Increasing it raises material-axis 3D-US up to $\alpha{=}2$ and steadily lowers object-axis 3D-US (Table~\ref{tab:ablation-stage-full}), and overall 3D-US peaks at $\alpha{=}1.2$. Combining $\alpha{=}1.5$ with $r{=}32$ gives 66.2, no better than $\alpha{=}1.2$ alone, so Table~\ref{tab:3dus} reports \our{} with $\alpha{=}1.2$ and all other hyperparameters at their defaults.

The baselines are tuned with the same one-at-a-time protocol, followed by a combined run of their individually best values (Appendix~\ref{sec:baseline-tuning}). Across single-parameter changes, overall 3D-US ranges from 0.0 to 52.5 for OCE and from 20.5 to 34.7 for UCE. Table~\ref{tab:3dus} reports each baseline at this combined configuration, with UCE editing both the key and the value projections as in its original implementation (Table~\ref{tab:ablation-kv}). Across all configurations of the three sweeps, \our{} matches or exceeds in both forgetting and preservation every baseline configuration with mean forgetting above 0.15 (Figure~\ref{fig:pareto}).

\begin{table}[H]
\centering
\small
\caption{\our{} hyperparameter ablation: per-axis and overall 3D-US for the best
value of each hyperparameter when it alone is changed from the default ($r{=}16$, $\beta{=}1$, null-prompt
multiplier $5$, $\gamma{=}1$, $\lambda{=}1$, $\alpha{=}1$). The full sweep is in
Table~\ref{tab:ablation-stage-full}. The $\alpha{=}1.2$ row
is the configuration reported in Table~\ref{tab:3dus}.}
\label{tab:ablation-stage}
\begin{tabular}{lcccc}
\toprule
Configuration & Shape & Material & Object & Overall \\
\midrule
Default & 64.9 & 44.9 & 74.5 & 60.1 \\
\midrule
$r=32$ & 68.8 & 47.0 & 77.8 & 63.1 \\
$\beta=3$ & 61.8 & 43.9 & 83.6 & 61.0 \\
null-prompt mult.\ $=7.5$ & 65.9 & 45.5 & 74.8 & 60.8 \\
$\gamma=0$ & 65.3 & 44.9 & 74.3 & 60.2 \\
$\lambda=0.1$ & 67.2 & 44.9 & 74.8 & 60.9 \\
$\alpha=1.2$ & 77.7 & 56.4 & 67.7 & \textbf{66.7} \\
\bottomrule
\end{tabular}
\end{table}

\section{Conclusion}

We introduced \our{}, a training-free, closed-form framework for
stage-aware concept erasure in native text-to-3D generation.
Our analysis on TRELLIS shows that the stage an edit must reach
depends on concept type. By combining a subspace derived from
the differences between erase and anchor embeddings with an affine correction,
\our{} enables targeted weight edits without additional inference
cost. Across 15 shape, material, and object concepts, it achieves
a 3D Unlearning Score of 66.7, compared with 53.2 for the strongest
adapted baseline under the same stage assignment. These results
support stage-aware editing as an effective approach to balancing
concept removal and preservation in native 3D generation.

\paragraph{Limitations}
The scope of quantitative erasure evaluation depends on the pretrained generator's capabilities and the available evaluators. Our protocol therefore verifies that target concepts
are expressed and recognized before editing, and combines semantic and physical evidence to assess their removal. Extending this evaluation to finer-grained and domain-specific
concepts requires distinguishing the removal of a target attribute from changes to closely related attributes. Developing benchmarks that capture these distinctions is a natural direction for future work.



\bibliographystyle{iclr2027_conference}

\clearpage
\appendix
\section{Concepts and Prompt Construction}
\label{sec:app-data}

This section describes the concepts we evaluate on, the prompts used to fit each edit, and the prompts used to evaluate it, in enough detail to reproduce the evaluation or extend it to new concepts.

\paragraph{Concepts and Anchors}
We evaluate on 15 concepts, five on each of three axes. Every concept is paired with a safe anchor from the same axis, so an edit replaces the concept with a plausible alternative of the same kind instead of an unrelated one. Table~\ref{tab:app-concepts} lists each anchor together with the label under which every evaluator recognizes the concept (Appendix~\ref{sec:app-metrics}).

\begin{table}[H]
\centering
\small
\caption{Concepts, anchors, and evaluator labels. Each shape concept is named in the editing prompts by the two words given in parentheses. The evaluator label is the text used by CLIP and Uni3D, and for materials the MINC-23 class is the material name itself.}
\label{tab:app-concepts}
\begin{tabular}{lllll}
\toprule
Axis & Concept & Anchor & Evaluator label & Evaluators \\
\midrule
Shape & sphere (\emph{round}, \emph{circular}) & cylinder & ``Object of sphere shape'' & CLIP, Uni3D \\
      & cube (\emph{cubic}, \emph{square}) & sphere & ``Object of cube shape'' & \\
      & cylinder (\emph{cylinder}, \emph{cylindrical}) & torus & ``Object of cylinder shape'' & \\
      & cellular lattice (\emph{cellular}, \emph{lattice}) & torus & ``Object of cellular shape'' & \\
      & torus (\emph{ring}, \emph{torus}) & cylinder & ``Object of torus shape'' & \\
\midrule
Material & wood & metal & ``Object made of wood'' & CLIP, MINC-23 \\
         & metal & wood & ``Object made of metal'' & \\
         & glass & ceramic & ``Object made of glass'' & \\
         & stone & ceramic & ``Object made of stone'' & \\
         & ceramic & glass & ``Object made of ceramic'' & \\
\midrule
Object & car & cat & ``A car'' & CLIP, Uni3D \\
       & cat & chair & ``A cat'' & \\
       & chair & teddy bear & ``A chair'' & \\
       & table & car & ``A table'' & \\
       & teddy bear & table & ``A teddy bear'' & \\
\bottomrule
\end{tabular}
\end{table}

\paragraph{Editing Prompts}
Each concept has 120 erase prompts and 120 anchor prompts, and the two sets are template-parallel. Erase prompt $i$ and anchor prompt $i$ come from the same sentence template and differ only in the concept word, as in \emph{``An analog clock crafted from wood.''} and \emph{``An analog clock crafted from metal.''}. In shape and material prompts, the concept describes an everyday object such as a cushion, a lamp, a vase, or a clock, and the templates vary how it is attached to that object, as in \emph{``An analog clock is strictly circular.''} or \emph{``A soft cushion made of wood.''}. All shape concepts share the same 120 templates, and so do all material concepts. In object prompts, the concept is the object itself. The object concepts share the same 120 sentence frames, with some adjectives chosen to suit each object, as in \emph{``A fast and sleek car.''} and \emph{``A small and fluffy cat.''}. Each shape concept is named by two different words, such as \emph{round} and \emph{circular} for sphere (Table~\ref{tab:app-concepts}), and each word appears in 60 of its 120 erase prompts, so the edit removes the shape under both names. Because the anchor word replaces the concept word in the same template, some anchor prompts are not fully grammatical, for example \emph{``An analog clock is strictly ring.''}. This is intended, since each pair then differs only in the concept word. When the two words differ in token count, as for \emph{chair} and \emph{teddy bear}, \our{} pairs only the final positions, where both states already encode the full prompt (Appendix~\ref{sec:app-impl}). The same 120 pairs are given to all three editors, and none of them is used for evaluation. Table~\ref{tab:app-prompts} shows one example per axis.

\paragraph{Evaluation Prompts}
Evaluation prompts are longer descriptions of a single asset that clearly shows the concept (Table~\ref{tab:app-prompts}). To stay close to the prompts TRELLIS handles best, we first studied the TRELLIS paper~\citep{xiang2025trellis} and the text prompts behind the example assets shown on its official project page, and followed their style as closely as possible. For each concept, the Gemini language model then generated a pool of about 200 candidate prompts in this style. We run a seed search over these candidates on the unedited TRELLIS. For each candidate in turn, we generate the asset with seeds $0,1,\dots,9$ and accept the first (prompt, seed) pair whose asset every evaluator of the axis assigns to the target concept, using one rendered view for CLIP and MINC-23 and the point cloud for Uni3D. The search stops once 30 pairs are accepted, which every concept reached. The accepted pairs are stored together with their seeds, and every method and ablation regenerates exactly these 30 pairs before and after its edit. This keeps only prompts on which TRELLIS expresses the concept at all, since forgetting cannot be measured otherwise. At the concept level, 3D-US additionally requires $\hat E^{\mathrm{pre}}_{e,c}\ge\tau=0.5$ for every evaluator on four views (Section~\ref{sec:3dus}), which all 15 concepts satisfy.

\paragraph{Caption Corpus for $K_0$}
The second moment $K_0$ of Section~\ref{sec:mods} is computed once from 30{,}000 captions of the ObjaverseXL (Sketchfab) part of TRELLIS-500K~\citep{xiang2025trellis}. The captions are encoded with the CLIP ViT-L/14 text encoder that conditions TRELLIS, and $K_0$ is the mean outer product of all non-padding token states. The same $K_0$ is given to all three editors.

\paragraph{Adding a Concept}
A new concept is added in four steps. We first choose an anchor on the same axis and the evaluator label of the concept. The erase prompts are then written by inserting the concept word into the templates of the axis, and the anchor prompts by inserting the anchor word into the same templates. Candidate evaluation prompts are collected and passed through the seed search above until 30 (prompt, seed) pairs are accepted. Finally, the concept enters the evaluation only if $\hat E^{\mathrm{pre}}_{e,c}\ge\tau$ holds for every evaluator of its axis.

\begin{table}[H]
\centering
\small
\caption{Example prompts for one concept per axis. Erase and anchor prompts fit the edit. Evaluation prompts are separate descriptive prompts, each fixed to a seed.}
\label{tab:app-prompts}
\begin{tabular}{lp{2.8cm}p{2.8cm}p{4.6cm}}
\toprule
Concept & Erase prompt & Anchor prompt & Evaluation prompt (seed) \\
\midrule
cylinder & An analog clock is strictly cylindrical. & An analog clock is strictly ring. & A portable audio speaker designed as a sleek, perfectly uniform cylindrical tube. (0) \\
\midrule
wood & An analog clock crafted from wood. & An analog clock crafted from metal. & A heavy medieval-style banquet table constructed from dark walnut with thick, hand-hewn wooden planks. (1) \\
\midrule
cat & A small and fluffy cat. & A small and fluffy chair. & A stretching tabby cat arching its back and extending its claws. (0) \\
\bottomrule
\end{tabular}
\end{table}

\section{Implementation of the Editors}
\label{sec:app-impl}

\paragraph{Shared Pipeline}
All three editors are closed-form and run through one shared pipeline, so they differ only in the update they compute. They receive the same 120 erase and anchor prompt pairs per concept (Appendix~\ref{sec:app-data}) and no additional preservation prompts. All prompts are encoded once with the frozen CLIP ViT-L/14 text encoder that conditions TRELLIS, with 77 token positions. UCE and OCE use one embedding per prompt, the hidden state at its last content token, and \our{} uses the aligned token states of Section~\ref{sec:mods}, which Table~\ref{tab:ablation-components} compares with the single embedding per prompt. All editors receive the same caption second moment $K_0$. Every editor modifies the text-conditioning projection (\texttt{to\_kv}, which maps the text states to keys and values) of every cross-attention layer in the stages selected by the per-axis mode of Section~\ref{sec:method-modes}, which is 28 layers per stage. \our{} and UCE edit both the key and the value half of each projection, as the original UCE does. OCE edits the value half, which at its tuned configuration scores far higher here than the key half edited by the original OCE or both halves (Table~\ref{tab:ablation-kv}). After the edit, all methods generate the same (prompt, seed) pairs with the same sampler settings and are scored by the same evaluation code (Appendix~\ref{sec:app-metrics}). The hyperparameters of every method are selected with the same one-at-a-time protocol (Appendix~\ref{sec:baseline-tuning}).

\paragraph{UCE}
We implement UCE~\citep{gandikota2024uce} with the closed-form update of the original paper and its official code. For every edited projection $W$, UCE solves a regularized least-squares problem that maps the erase embeddings $c_1,\dots,c_k$ to the output of a guide concept and keeps the output unchanged on a preserve set $\mathcal{P}$:
\begin{equation*}
\min_{W'}\ \sum_{i=1}^{k}\big\lVert W'c_i - W\bar c^*\big\rVert_2^2
\;+\;\frac{\gamma_{\mathrm{g}}}{|\mathcal{P}|}\sum_{c\in\mathcal{P}}\big\lVert (W'-W)c\big\rVert_2^2
\;+\;\lambda\,\lVert W'-W\rVert_F^2 .
\end{equation*}
As in the original, the update is applied to both the key and the value projections, each prompt is represented by the text-encoder state at its last content token, and the ridge term $\lambda$ is the regularizer of the official implementation. Our setting changes two of its inputs. First, the guide is the mean anchor embedding $\bar c^*$, so every erase prompt of a concept is redirected to one target, as when UCE is given a single guide concept. Second, the original preserves an explicit list of concepts, and no editor receives such a list here. We therefore take $\mathcal{P}$ to be the token states of the caption corpus of Appendix~\ref{sec:app-data}, which turns the preservation term into $\gamma_{\mathrm{g}}\,\mathrm{tr}\big((W'-W)K_0(W'-W)^\top\big)$ and gives UCE the same corpus statistic as \our{} and OCE. With $C=[c_1\cdots c_k]$, the minimizer is
$W' = W\big(\bar c^*\mathbf{1}^\top C^\top + \gamma_{\mathrm{g}}K_0 + \lambda I\big)\big(CC^\top + \gamma_{\mathrm{g}}K_0 + \lambda I\big)^{-1}$.
The swept hyperparameters are the preserve-global weight $\gamma_{\mathrm{g}}$ and the ridge weight $\lambda$ (Table~\ref{tab:ablation-uce}).

\paragraph{OCE}
We implement OCE~\citep{sun2026oce} with the rotation update of the original paper. Let $G$ and $G^*$ be the orthogonal projectors onto the spans of the erase images $\{Wc_i\}$ and the anchor images $\{Wc^*_i\}$. OCE seeks the rotation that maps the erase images onto the anchor images while preserving the outputs of other concepts, and solves it in a subspace form. The original objective matrix is
$-s_{\mathrm{e}}\,G(I-G^*) + W\big(\gamma_{\mathrm{g}}K_0 + \lambda_{\mathrm{r}}C_nC_n^\top\big)W^\top$,
whose last term preserves an explicit list of neighboring concepts $C_n$. Our setting changes two of its inputs. As for the other editors, no such list is given, so the $\lambda_{\mathrm{r}}$ term is not used, and $K_0$ is computed on the TRELLIS caption corpus rather than on the photo captions used for 2D image diffusion, since TRELLIS is conditioned on a different caption distribution. We also add a ridge term $\lambda\,WW^\top$, which is the same as adding $(\lambda/\gamma_{\mathrm{g}})\,I$ to $K_0$ and plays the role of the ridge weight of UCE. Setting $\lambda=0$ recovers the original objective, which reaches an overall 3D-US of 52.1, close to the 53.2 of Table~\ref{tab:3dus} (Table~\ref{tab:ablation-kv}). With the ridge term, the objective matrix becomes
\begin{equation*}
M = -s_{\mathrm{e}}\,G\,(I-G^*) \;+\; \gamma_{\mathrm{g}}\,WK_0W^\top \;+\; \lambda\,WW^\top .
\end{equation*}
As in the original, OCE takes the orthogonal matrix closest to $M$, $R=UV^\top$ from the SVD $M=U\Sigma V^\top$ (flipping the sign of the last column of $U$ if needed so that $\det R=1$), and sets $W'=RW$. The $K_0$ and ridge terms are positive semidefinite, and the rotation closest to a positive semidefinite matrix is the identity, so these terms keep $R$ close to the identity while the erase term drives the change. The swept hyperparameters are the erase scale $s_{\mathrm{e}}$ and the preserve-global weight $\gamma_{\mathrm{g}}$, written $\lambda_{\mathrm{e}}$ and $\lambda_0$ in the original paper, and $\lambda$ (Table~\ref{tab:ablation-oce}).

\paragraph{\our{}}
\our{} follows Section~\ref{sec:mods}. For a template-parallel pair whose prompts have the same token count, all 77 aligned positions are paired, and positions whose two states coincide (relative difference below $10^{-4}$) are dropped. These are the positions before the concept word, which are identical under the causal text encoder. When the token counts differ, positions from the later of the two last content tokens onward are paired, where both states already encode the full prompt. This gives between 8{,}317 and 8{,}627 token pairs per concept. Pairing every position matches how TRELLIS reads the text, since it feeds all 77 token states, including those after the end-of-text token, to every cross-attention layer without a padding mask. The text embedding also reaches the \texttt{to\_kv} projections without any normalization, so folding $T$ into both stages (Mode-SL) is equivalent to replacing every text embedding $c$ by $T(c)$, and Mode-S or Mode-L applies this replacement for one stage only. The retained set $\mathcal{H}$ contains the anchor state of every pair, with weight $\beta$, and all 77 token states of the empty prompt, which TRELLIS uses as the unconditional branch of classifier-free guidance, with weight $\beta$ times the null-prompt multiplier. The subspace rank is capped at $r$ and at singular values below $10^{-3}$ of the largest one. All solves use double precision, and the correction is folded into both the weight and the bias of each projection (equation~(\ref{eq:stage_fold})). Table~\ref{tab:ablation-stage-full} lists the swept values of $r$, $\beta$, the null-prompt multiplier, $\gamma$, $\lambda$, and $\alpha$.

\section{Evaluation Protocol}
\label{sec:app-metrics}

\paragraph{Renders and Point Clouds}
Each generated asset is rendered from the Gaussian output of TRELLIS at $N_v=4$ camera positions spread over the sphere by a golden-angle (Fibonacci) spiral, at radius 2 with a $40^\circ$ field of view, at $512\times512$ resolution with $4\times$ supersampling on a black background. The point cloud for Uni3D uses the centers of the Gaussians with opacity above $0.1$, colored by their zeroth-order spherical-harmonic coefficient. We sample 10{,}000 points with a fixed random seed and normalize them to zero mean and unit radius.

\paragraph{Evaluators}
The semantic evaluator is OpenCLIP ViT-B/32 trained on LAION-2B (\texttt{laion/CLIP-ViT-B-32-laion2B-s34B-b79K}). It is a different model from the CLIP ViT-L/14 encoder that conditions TRELLIS, so the evaluator does not share weights with the representation being edited. For shape and object concepts, the physical evaluator is Uni3D~\citep{zhou2024uni3d} with its largest point-cloud backbone (Uni3D-g) and the EVA02-E/14+ text tower it was aligned with, used zero-shot. For material concepts, the physical evaluator is a SigLIP image classifier fine-tuned on the 23 MINC categories~\citep{bell2015minc} (\texttt{prithivMLmods/Minc-Materials-23}), whose output is restricted to the five materials of the axis and renormalized. The restriction gives all evaluators the same closed candidate set $\mathcal{C}$ and the same chance level.

\paragraph{Detection and Alignment}
For detection, CLIP and MINC-23 predict the top-1 label among the five labels of the axis for every view, and a concept counts as detected in an asset if it wins at least one view. Uni3D makes one prediction per point cloud. $E$ is the percentage of the 30 evaluation assets in which the concept is detected, and the chance level is $\rho_e=20$. For preservation, the CLIP alignment $A_e(x,c')$ is the cosine similarity between the rendered views of $x$ and the text of the prompt that generated $x$, averaged over views, and $\bar A_e(x,c')$ is the mean similarity of the same views to the labels of the other four concepts of the axis. Uni3D uses the same definition with the point cloud in place of the views. For MINC-23, $\hat A$ is the chance-corrected probability of the asset's own material under the restricted classifier, averaged over views.

\paragraph{Agreement Between Evaluators}
Each axis is scored by two evaluators that see different evidence, rendered images and either a point cloud or a material classifier. We measure how often they agree on the top-1 label of the same asset (Table~\ref{tab:app-agreement}), using the runs of Table~\ref{tab:3dus}. Before the edit, agreement is high for shape and object concepts. After the edit, the generations cover everything from unchanged assets to fully redirected ones and differ between methods. On these, the two evaluators agree on 69 to 74\% of the assets, with Cohen's $\kappa$ between $0.60$ and $0.67$, far above the value of $0$ expected by chance. The image and 3D evaluators therefore track the same change, while each also registers changes the other does not, which is what the geometric mean of Section~\ref{sec:3dus} combines. The ranking of Table~\ref{tab:3dus} also does not depend on how 3D-US aggregates its parts. \our{} ranks first, ahead of OCE and UCE by at least 11 points, when the geometric mean over axes is replaced by the arithmetic mean, when the evaluators are combined by their arithmetic mean or their minimum instead of the geometric mean, and when the mean $F$ and $P$ over all 15 concepts are combined by their harmonic or arithmetic mean.

\begin{table}[H]
\centering
\small
\caption{Agreement on the top-1 label between the two evaluators of each axis, with Cohen's $\kappa$. Image evaluators use the view-averaged prediction. Each of the 15 runs per axis (three methods, five erased concepts) generates the 150 evaluation assets of the axis before and after its edit, which gives 2{,}250 assets per axis in each column.}
\label{tab:app-agreement}
\begin{tabular}{llcccc}
\toprule
 & & \multicolumn{2}{c}{Before edit} & \multicolumn{2}{c}{After edit} \\
\cmidrule(lr){3-4}\cmidrule(lr){5-6}
Axis & Evaluators & Agreement & $\kappa$ & Agreement & $\kappa$ \\
\midrule
Shape & CLIP, Uni3D & 97.6\% & 0.969 & 73.9\% & 0.671 \\
Material & CLIP, MINC-23 & 77.3\% & 0.715 & 68.8\% & 0.600 \\
Object & CLIP, Uni3D & 98.6\% & 0.983 & 72.2\% & 0.651 \\
\bottomrule
\end{tabular}
\end{table}

\section{Detailed Results}
\label{sec:app-perconcept}

This section breaks the comparison of Table~\ref{tab:3dus} down by concept and checks what an erased concept is replaced with. All values come from the same runs as Table~\ref{tab:3dus}.

\paragraph{Per-Concept Scores}
Table~\ref{tab:app-perconcept} lists forgetting, preservation, and 3D-US for every concept. Their means over the five concepts of an axis are the axis entries of Table~\ref{tab:3dus}. \our{} has the highest 3D-US on 10 of the 15 concepts.

\begin{table}[H]
\centering
\small
\setlength{\tabcolsep}{4pt}
\caption{Per-concept forgetting ($F$), preservation ($P$), and 3D-US for the runs of Table~\ref{tab:3dus}. Bold marks the highest 3D-US for each concept.}
\label{tab:app-perconcept}
\begin{tabular}{ll ccc ccc ccc}
\toprule
 & & \multicolumn{3}{c}{UCE} & \multicolumn{3}{c}{OCE} & \multicolumn{3}{c}{\our{}} \\
\cmidrule(lr){3-5}\cmidrule(lr){6-8}\cmidrule(lr){9-11}
Axis & Concept & $F$ & $P$ & 3D-US & $F$ & $P$ & 3D-US & $F$ & $P$ & 3D-US \\
\midrule
Shape & Sphere & 0.63 & 0.83 & 71.7 & 0.96 & 0.57 & 71.5 & 0.63 & 0.94 & \textbf{75.2} \\
 & Cube & 1.00 & 0.82 & 90.1 & 0.71 & 0.69 & 69.7 & 1.00 & 0.89 & \textbf{94.1} \\
 & Cylinder & 0.14 & 0.82 & 24.6 & 0.43 & 0.68 & 52.8 & 0.45 & 0.87 & \textbf{59.0} \\
 & Cellular lattice & 0.70 & 0.87 & \textbf{77.4} & 0.54 & 0.69 & 60.6 & 0.67 & 0.90 & 76.4 \\
 & Torus & 0.58 & 0.91 & 70.8 & 1.00 & 0.73 & \textbf{84.7} & 0.75 & 0.95 & 83.6 \\
\midrule
Material & Wood & 0.08 & 0.83 & 15.1 & 0.24 & 0.71 & 35.8 & 0.60 & 0.97 & \textbf{74.3} \\
 & Metal & 0.19 & 0.77 & 30.6 & 0.29 & 0.75 & 41.7 & 0.36 & 0.85 & \textbf{50.6} \\
 & Glass & 0.27 & 0.85 & 41.0 & 0.35 & 0.72 & 47.5 & 0.58 & 0.91 & \textbf{71.0} \\
 & Stone & 0.54 & 0.89 & 67.3 & 0.92 & 0.77 & \textbf{83.8} & 0.67 & 0.94 & 78.1 \\
 & Ceramic & 0.00 & 0.88 & 0.0 & 0.00 & 0.73 & 0.0 & 0.04 & 0.94 & \textbf{8.0} \\
\midrule
Object & Car & 0.00 & 0.67 & 0.0 & 0.41 & 0.47 & 43.8 & 0.71 & 0.91 & \textbf{79.5} \\
 & Cat & 0.84 & 0.59 & \textbf{69.6} & 1.00 & 0.44 & 61.1 & 0.79 & 0.22 & 34.5 \\
 & Chair & 0.47 & 0.67 & 55.1 & 0.74 & 0.38 & 50.5 & 1.00 & 0.85 & \textbf{91.7} \\
 & Table & 0.52 & 0.74 & 61.0 & 1.00 & 0.35 & 51.6 & 1.00 & 0.83 & \textbf{90.7} \\
 & Teddy bear & 1.00 & 0.58 & \textbf{73.8} & 1.00 & 0.42 & 58.9 & 1.00 & 0.27 & 42.2 \\
\bottomrule
\end{tabular}
\end{table}

\begin{table}[H]
\centering
\small
\caption{Redirect rate (\%), the share of edited evaluation assets of the erased concept that each evaluator assigns to the anchor, averaged over the five concepts of each axis. Chance is 20\%.}
\label{tab:app-redirect}
\begin{tabular}{llccc}
\toprule
Axis & Evaluator & UCE & OCE & \our{} \\
\midrule
Shape & CLIP & 47.3 & 28.0 & \textbf{51.3} \\
      & Uni3D & 38.0 & 28.0 & \textbf{42.7} \\
\midrule
Material & CLIP & 20.7 & 22.7 & \textbf{39.3} \\
         & MINC-23 & 22.7 & 18.0 & \textbf{32.7} \\
\midrule
Object & CLIP & 34.0 & 41.3 & \textbf{83.3} \\
       & Uni3D & 39.3 & 42.0 & \textbf{80.0} \\
\midrule
All & mean & 33.7 & 30.0 & \textbf{54.9} \\
\bottomrule
\end{tabular}
\end{table}

\paragraph{Redirection to the Anchor}
Forgetting counts an erased concept as removed whatever replaces it. To check that the edits produce the intended replacement, we also measure the redirect rate, the percentage of the evaluation assets of the erased concept whose top-1 label after the edit is its anchor from Table~\ref{tab:app-concepts}. The image evaluators use the view-averaged prediction, as in Table~\ref{tab:app-agreement}, so that every asset receives exactly one label, like the single prediction of Uni3D on the point cloud. The any-view rule of forgetting suits detection, where a concept visible from one view is still present, but can assign several labels to the same asset. With five candidate labels, chance is 20\%. Table~\ref{tab:app-redirect} shows that \our{} reaches the anchor far more often than both baselines on every axis and under every evaluator.

\section{Ablations}
\label{sec:baseline-tuning}

This section gives the full hyperparameter sweeps of the three editors and the choice of edited projections. All three editors are tuned with the same one-at-a-time protocol. Starting from a default configuration, we change one hyperparameter at a time while the others keep their default values, and evaluate every configuration on all 15 concepts with the protocol of Section~\ref{sec:setup}.

\paragraph{\our{}}
Table~\ref{tab:ablation-stage-full} gives the full sweep summarized in Table~\ref{tab:ablation-stage}. The configuration reported in Table~\ref{tab:3dus} changes only the overshoot scale, to $\alpha{=}1.2$. Combining a larger overshoot with the best change of any other hyperparameter ($\alpha{=}1.5$, $r{=}32$) does not improve on it.

\begin{table}[H]
\centering
\small
\caption{Full \our{} hyperparameter sweep: per-axis and overall 3D-US when one
hyperparameter is changed from the default ($r{=}16$, $\beta{=}1$, null-prompt
multiplier $5$, $\gamma{=}1$, $\lambda{=}1$, $\alpha{=}1$). The $\alpha{=}1.2$ row
is the configuration reported in Table~\ref{tab:3dus}.}
\label{tab:ablation-stage-full}
\begin{tabular}{lcccc}
\toprule
Configuration & Shape & Material & Object & Overall \\
\midrule
Default & 64.9 & 44.9 & 74.5 & 60.1 \\
\midrule
$r=8$ & 67.0 & 45.4 & 71.8 & 60.2 \\
$r=32$ & 68.8 & 47.0 & 77.8 & 63.1 \\
\midrule
$\beta=0.3$ & 66.4 & 46.0 & 62.0 & 57.5 \\
$\beta=3$ & 61.8 & 43.9 & 83.6 & 61.0 \\
\midrule
null-prompt mult.\ $=1$ & 70.7 & 45.4 & 68.4 & 60.3 \\
null-prompt mult.\ $=7.5$ & 65.9 & 45.5 & 74.8 & 60.8 \\
\midrule
$\gamma=0$ & 65.3 & 44.9 & 74.3 & 60.2 \\
$\gamma=3$ & 64.6 & 44.5 & 73.7 & 59.6 \\
\midrule
$\lambda=0.1$ & 67.2 & 44.9 & 74.8 & 60.9 \\
$\lambda=10$ & 66.3 & 44.5 & 76.1 & 60.8 \\
\midrule
$\alpha=1.2$ & 77.7 & 56.4 & 67.7 & \textbf{66.7} \\
$\alpha=1.5$ & 78.0 & 63.7 & 58.5 & 66.3 \\
$\alpha=1.7$ & 73.3 & 65.8 & 53.5 & 63.7 \\
$\alpha=2.0$ & 67.7 & 69.9 & 39.4 & 57.1 \\
$\alpha=2.5$ & 45.4 & 69.6 & 22.8 & 41.6 \\
\midrule
Combined ($\alpha{=}1.5$, $r{=}32$) & 77.6 & 62.7 & 59.7 & 66.2 \\
\bottomrule
\end{tabular}
\end{table}

\paragraph{Components of \our{}}
Table~\ref{tab:ablation-components} removes or replaces one component of \our{} at a time, at the default configuration ($\alpha{=}1$). Without the offset $n_0$, the correction can only act through the subspace term, forgetting drops on every axis (mean $F$ of 0.31, 0.20, and 0.47 instead of 0.54, 0.33, and 0.89), and overall 3D-US falls to 41.3. Without the subspace confinement ($D{=}I$), the correction reads the whole embedding, which raises forgetting on the shape axis but lowers preservation on every axis, most on the object axis (mean $P$ of 0.36 instead of 0.73), and overall 3D-US falls to 54.1. Replacing the affine map with the orthogonal map inside the same subspace that best aligns erase and anchor states (orthogonal Procrustes, without offset) gives 45.3, which isolates the argument of Figure~\ref{fig:geometric_intuition}. Finally, fitting \our{} only on the last content token of each prompt, the single embedding per prompt that UCE and OCE receive (120 pairs instead of 8{,}317 to 8{,}627), gives 59.0, close to the 60.1 of token-level pairing and above both tuned baselines (53.2 for OCE and 47.5 for UCE). The advantage of \our{} therefore does not come from its additional token pairs.

\begin{table}[H]
\centering
\small
\caption{Components of \our{}: per-axis and overall 3D-US at the default configuration ($r{=}16$, $\beta{=}1$, null-prompt multiplier $5$, $\gamma{=}1$, $\lambda{=}1$, $\alpha{=}1$) when one component is removed or replaced. Last-token pairs fits \our{} on the same single embedding per prompt that UCE and OCE use.}
\label{tab:ablation-components}
\begin{tabular}{lcccc}
\toprule
Variant & Shape & Material & Object & Overall \\
\midrule
\our{} & 64.9 & 44.9 & 74.5 & \textbf{60.1} \\
\midrule
No subspace ($D{=}I$) & 69.3 & 43.5 & 52.4 & 54.1 \\
No offset ($n_0{=}0$) & 42.3 & 31.3 & 53.2 & 41.3 \\
Rotation in $V$ & 50.4 & 39.3 & 46.9 & 45.3 \\
Last-token pairs & 67.6 & 40.7 & 74.5 & 59.0 \\
\bottomrule
\end{tabular}
\end{table}

\paragraph{UCE and OCE}
Tables~\ref{tab:ablation-oce} and~\ref{tab:ablation-uce} report the sweeps of the two baselines. For each of them, ``Combined'' takes the individually best value of every hyperparameter that moved 3D-US by more than noise and evaluates these values together in a single run. The combined rows are the configurations reported in Table~\ref{tab:3dus}.

\begin{table}[H]
\centering
\small
\caption{OCE hyperparameter sweep: per-axis and overall 3D-US when one hyperparameter is changed from the default (erase scale ${=}2000$,
preserve-global ${=}10$, $\lambda{=}10$). The Combined row is the configuration reported in Table~\ref{tab:3dus}.}
\label{tab:ablation-oce}
\begin{tabular}{lcccc}
\toprule
Configuration & Shape & Material & Object & Overall \\
\midrule
Default & 35.4 & 25.0 & 56.7 & 36.9 \\
\midrule
erase scale $=500$ & 12.8 & 12.2 & 0.0 & 0.0 \\
erase scale $=5000$ & 58.4 & 32.1 & 66.2 & 49.9 \\
\midrule
preserve-global $=1$ & 65.0 & 34.2 & 65.3 & 52.5 \\
preserve-global $=50$ & 13.4 & 13.1 & 12.7 & 13.1 \\
\midrule
$\lambda=1$ & 39.3 & 28.1 & 63.6 & 41.3 \\
$\lambda=50$ & 21.8 & 18.8 & 19.6 & 20.0 \\
\midrule
Combined (erase${=}5000$, pres.-global${=}1$, $\lambda{=}1$) & 67.9 & 41.8 & 53.2 & \textbf{53.2} \\
\bottomrule
\end{tabular}
\end{table}

\begin{table}[H]
\centering
\small
\caption{UCE hyperparameter sweep: per-axis and overall 3D-US when one hyperparameter is changed from the default (preserve-global ${=}1$, $\lambda{=}1$). The Combined row is the configuration reported in Table~\ref{tab:3dus}.}
\label{tab:ablation-uce}
\begin{tabular}{lcccc}
\toprule
Configuration & Shape & Material & Object & Overall \\
\midrule
Default & 33.6 & 20.0 & 27.7 & 26.5 \\
\midrule
$\lambda=0.1$ & 24.1 & 15.3 & 23.4 & 20.5 \\
$\lambda=10$ & 42.6 & 23.1 & 31.8 & 31.5 \\
\midrule
preserve-global $=0.1$ & 48.3 & 24.4 & 35.4 & 34.7 \\
preserve-global $=5$ & 20.9 & 15.3 & 26.7 & 20.5 \\
\midrule
Combined ($\lambda{=}10$, preserve-global${=}0.1$) & 66.9 & 30.8 & 51.9 & \textbf{47.5} \\
\bottomrule
\end{tabular}
\end{table}

\paragraph{Edited Projections}
Each cross-attention layer reads the text through one projection whose output is split into keys and values. Table~\ref{tab:ablation-kv} compares editing the key half, the value half, or both. At the default configurations, all three editors do best when the value half is edited, alone or together with the keys, and editing only the keys erases almost nothing. \our{} edits both halves, since its correction is consistent for keys and values, and UCE edits both halves as in its original implementation. For OCE, the table also reports the configuration of Table~\ref{tab:3dus}, where the value edit scores far higher than the key edit of the original OCE or the edit of both halves. Setting $\lambda=0$ in the tuned OCE, which recovers its original objective, gives 52.1 instead of 53.2.

\begin{table}[H]
\centering
\small
\caption{Overall 3D-US when each editor modifies the key half (K), the value half (V), or both halves (K+V) of the cross-attention text projection, at the default configurations and, for OCE, at the configuration of Table~\ref{tab:3dus}. Bold marks the variant reported in Table~\ref{tab:3dus}.}
\label{tab:ablation-kv}
\begin{tabular}{llccc}
\toprule
Method & Configuration & K & V & K+V \\
\midrule
\our{} & default & 5.2 & 50.7 & 60.1 \\
OCE & default & 7.4 & 36.9 & 37.6 \\
UCE & default & 3.9 & 26.5 & 37.2 \\
\bottomrule
\end{tabular}
\end{table}

\begin{figure}[H]
\centering
\includegraphics[width=0.85\linewidth]{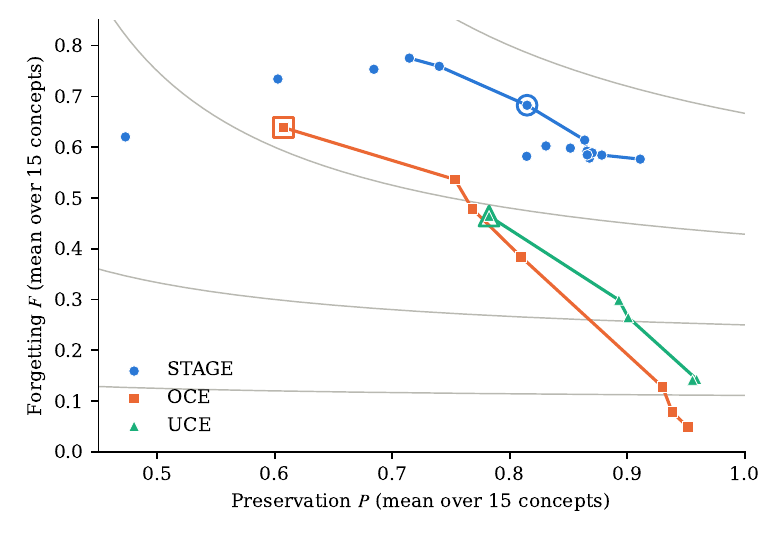}
\caption{Forgetting against preservation for the non-default configurations of Tables~\ref{tab:ablation-stage-full}, \ref{tab:ablation-oce}, and~\ref{tab:ablation-uce}, averaged over the 15 concepts. Lines connect the Pareto-optimal configurations of each method, open markers show the configurations of Table~\ref{tab:3dus}, and gray curves mark the levels 20, 40, 60, and 80 of $100\cdot 2FP/(F+P)$, rising toward the upper right.}
\label{fig:pareto}
\end{figure}

\paragraph{Forgetting and Preservation Trade-off}
Figure~\ref{fig:pareto} places every non-default configuration of the three sweeps by its mean forgetting and preservation over the 15 concepts. The configurations of \our{} form a front above those of both baselines. Every baseline configuration with mean forgetting above 0.15 is matched or exceeded in both forgetting and preservation by at least one configuration of \our{}. The baseline configurations outside this front forget almost nothing and keep high preservation only because they barely change the model.

\section{Qualitative Results}
\label{sec:app-qualitative}

\begin{figure}
    \centering
    \includegraphics[width=0.95\textwidth]{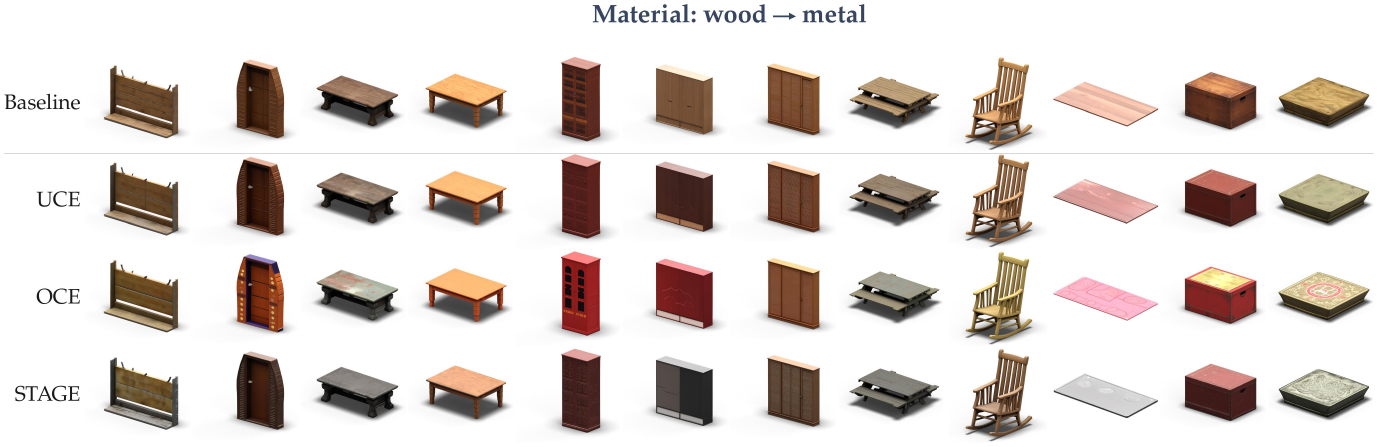}\\[0.5em]
    \includegraphics[width=0.95\textwidth]{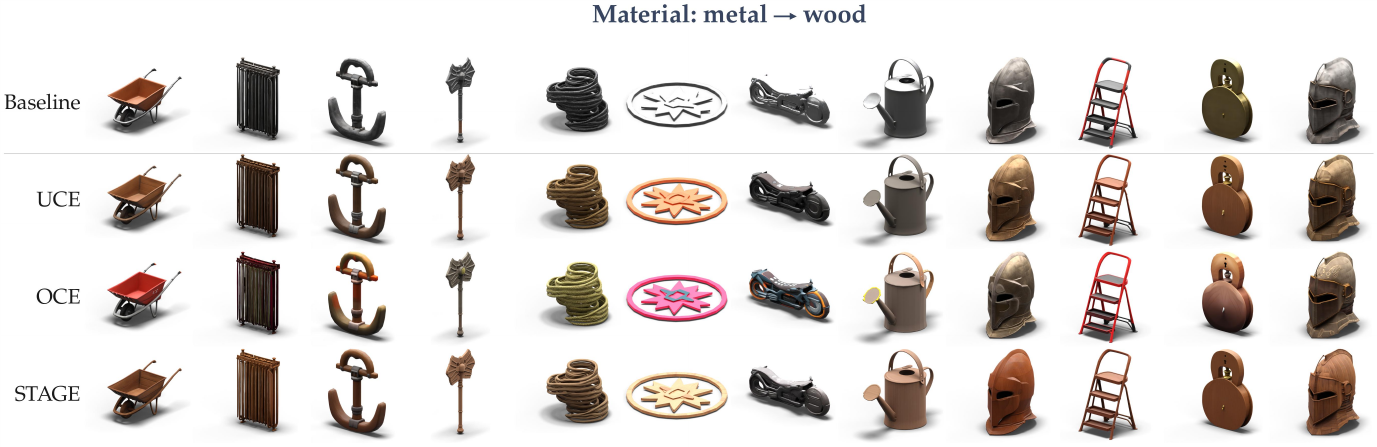}\\[0.5em]
    \includegraphics[width=0.95\textwidth]{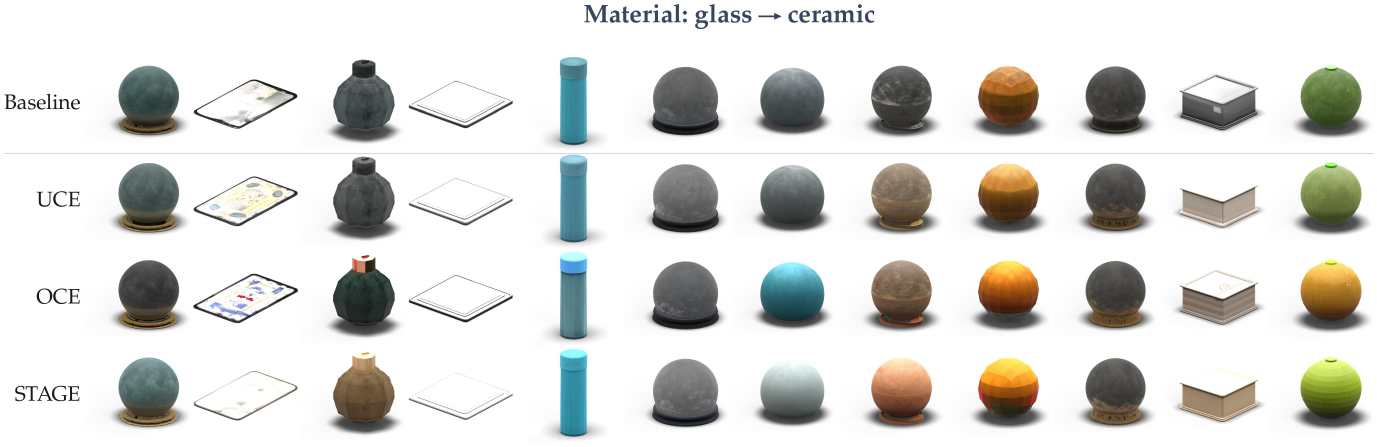}\\[0.5em]
    \includegraphics[width=0.95\textwidth]{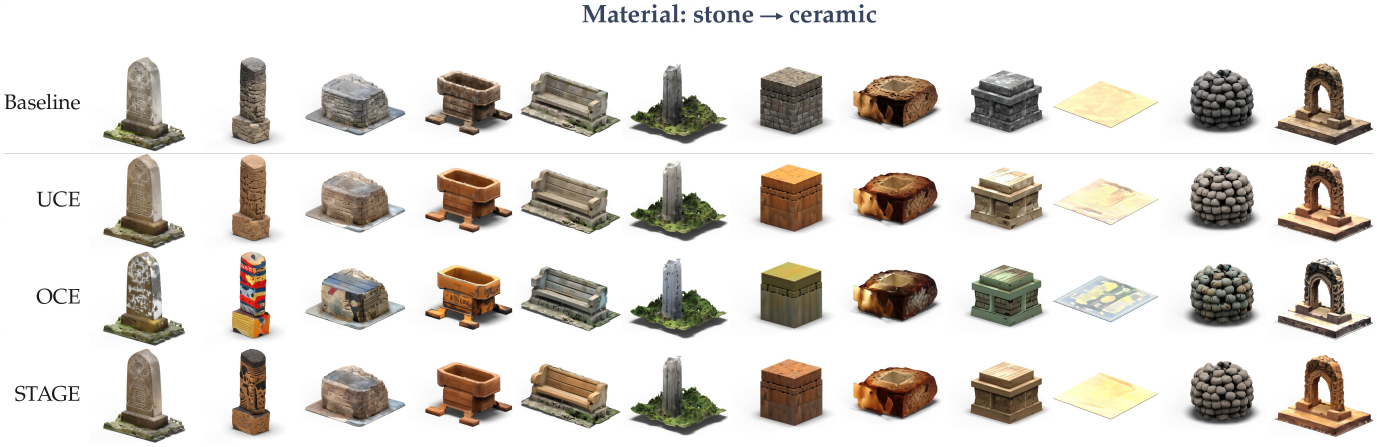}\\[0.5em]
    \includegraphics[width=0.95\textwidth]{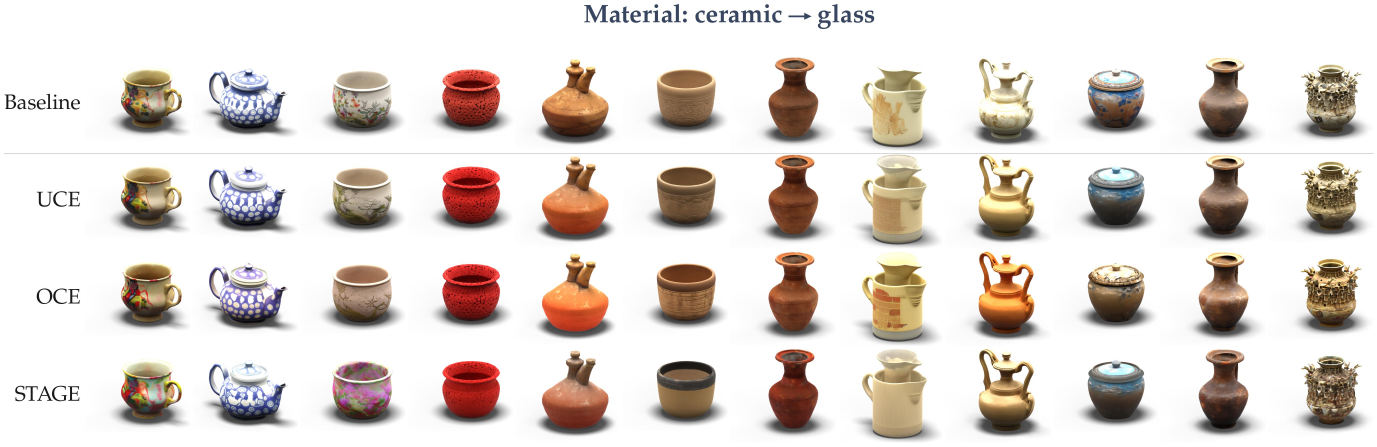}
    \caption{%
      Erasing material concepts. The \emph{Baseline} row shows the unedited model, and the rows below show the same twelve prompts after erasure with UCE, OCE, and \our{}. Models share a scale within a column, not across columns.%
    }
    \label{fig:strips_material}
  \end{figure}

\begin{figure}
      \centering
      \includegraphics[width=0.95\textwidth]{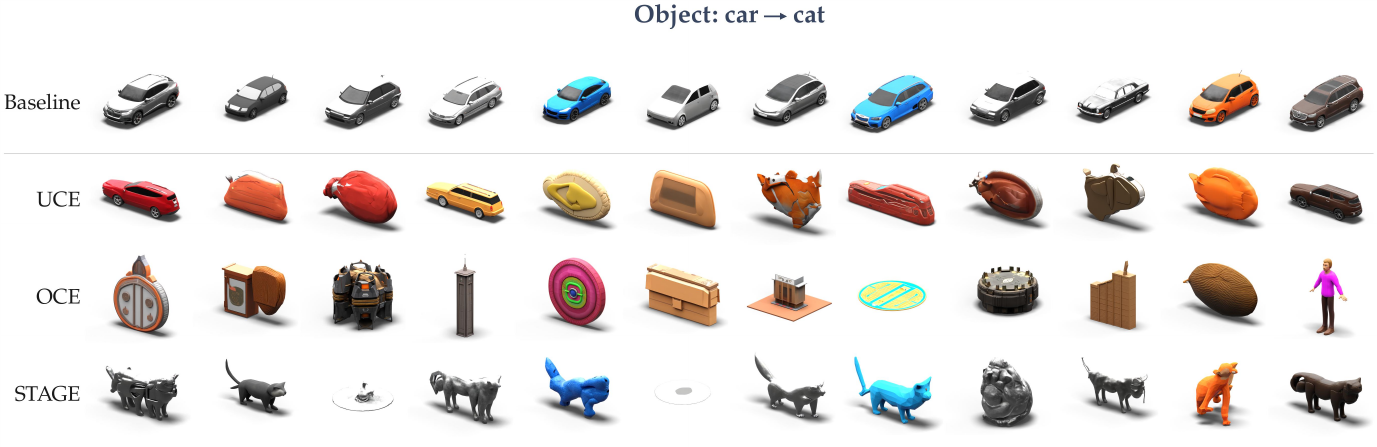}\\[0.5em]
      \includegraphics[width=0.95\textwidth]{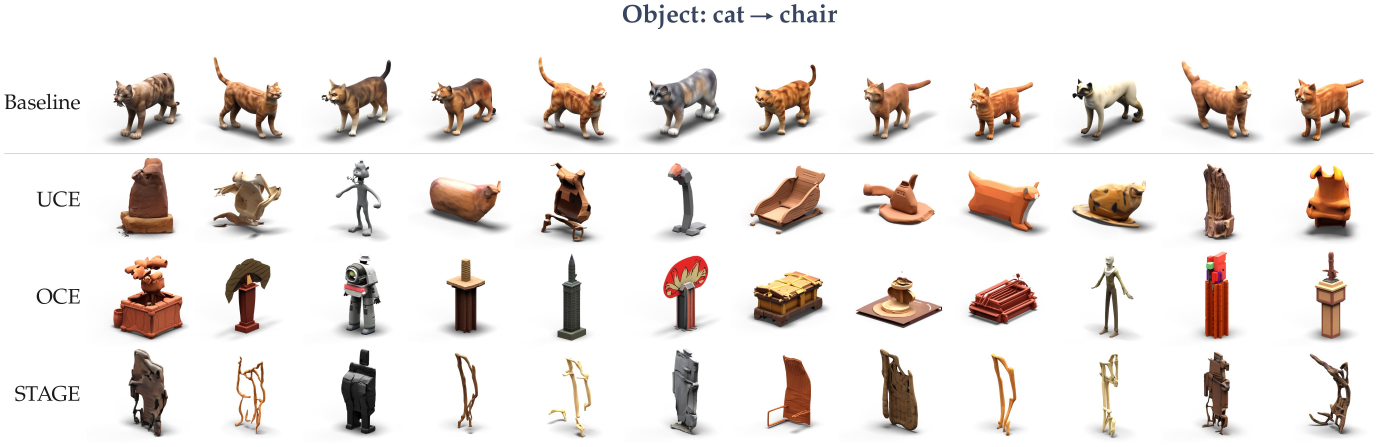}\\[0.5em]
      \includegraphics[width=0.95\textwidth]{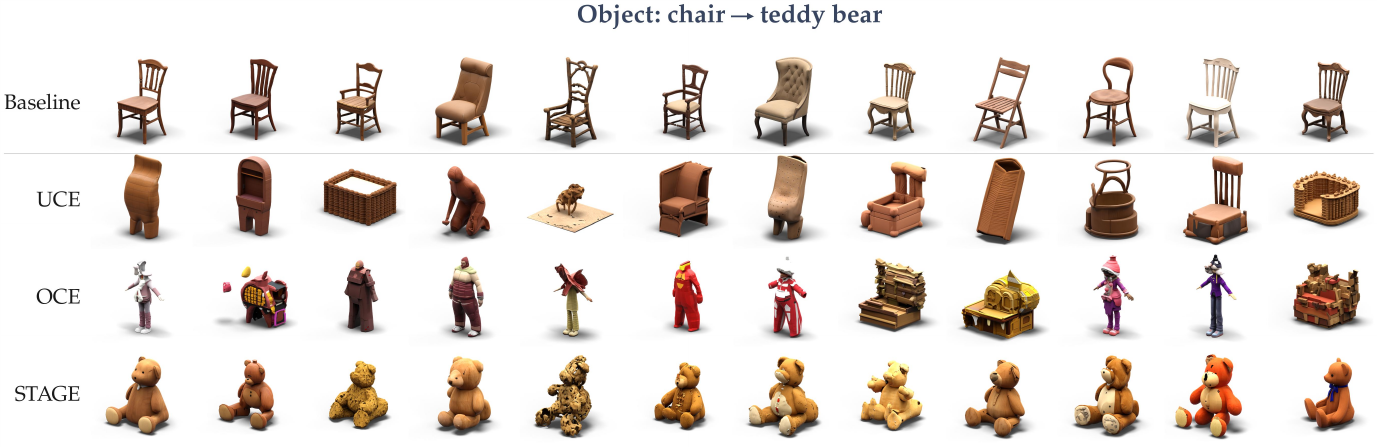}\\[0.5em]
      \includegraphics[width=0.95\textwidth]{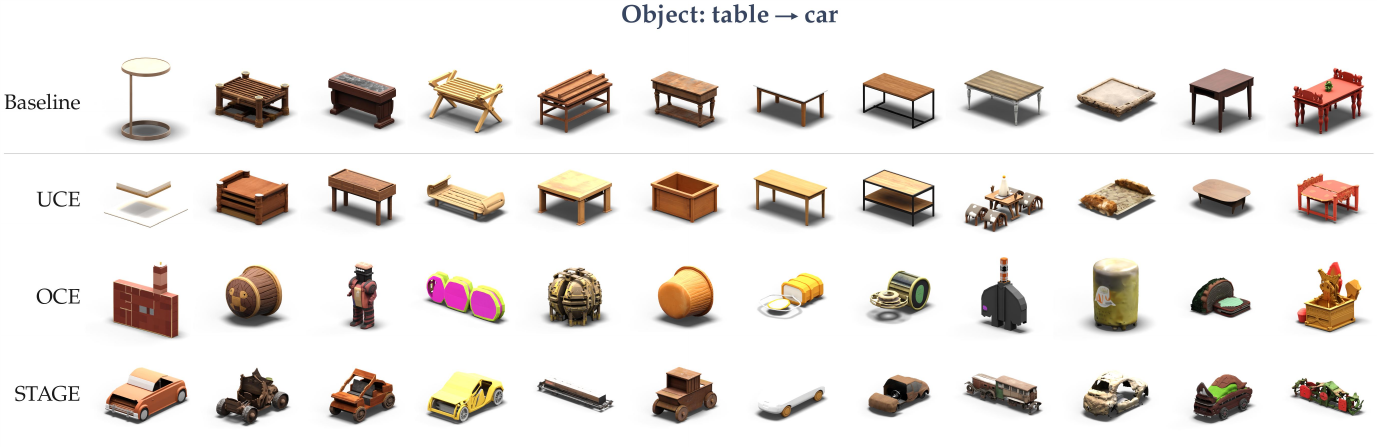}\\[0.5em]
      \includegraphics[width=0.95\textwidth]{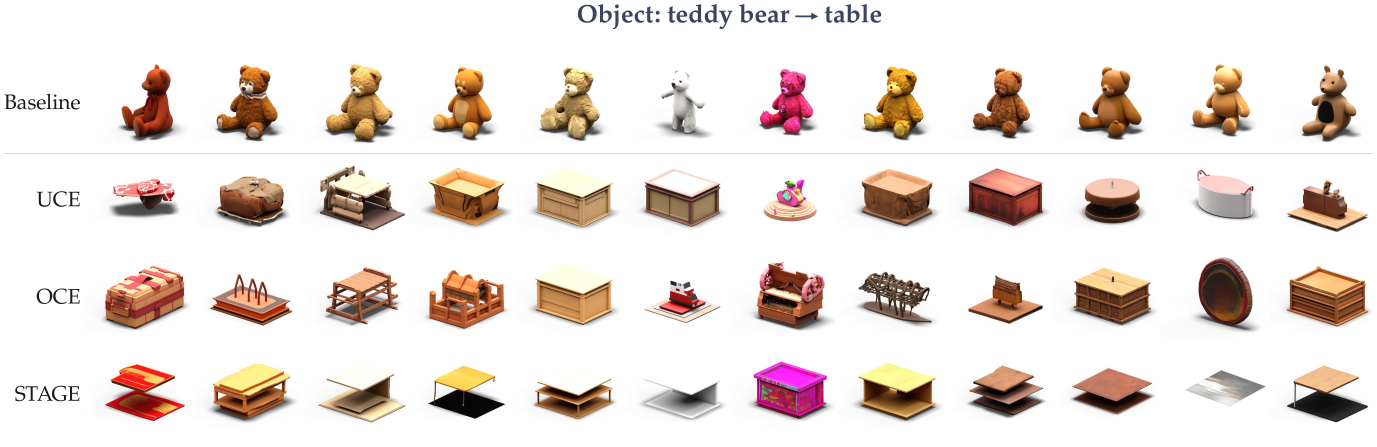}
      \caption{%
        Erasing object concepts. The \emph{Baseline} row shows the unedited model, and the rows below show the same twelve prompts after erasure with UCE, OCE, and \our{}. Models share a scale within a column, not across columns.%
      }
      \label{fig:strips_object}
    \end{figure}

\begin{figure}
      \centering
      \includegraphics[width=0.95\textwidth]{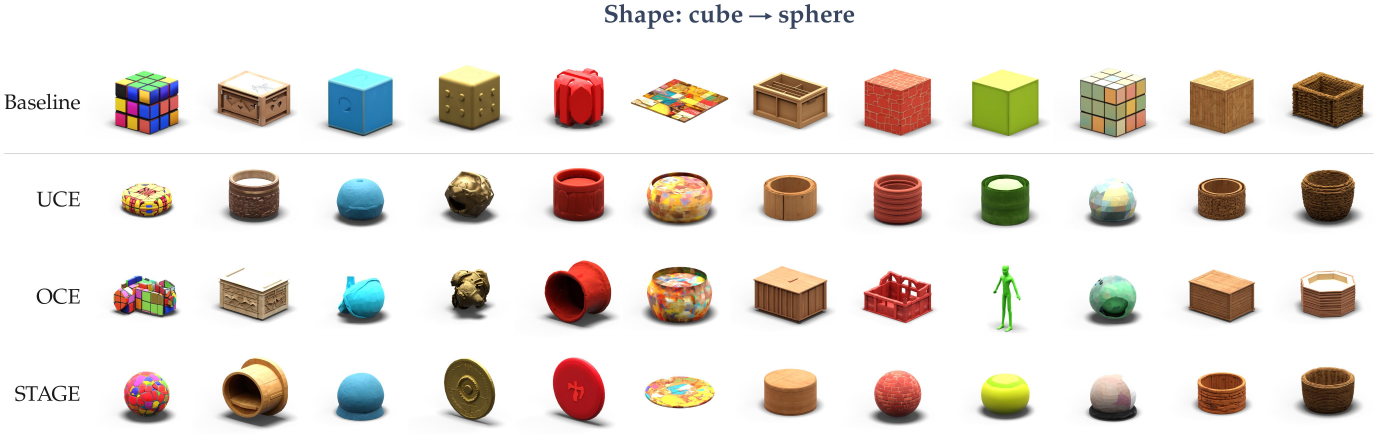}\\[0.5em]
      \includegraphics[width=0.95\textwidth]{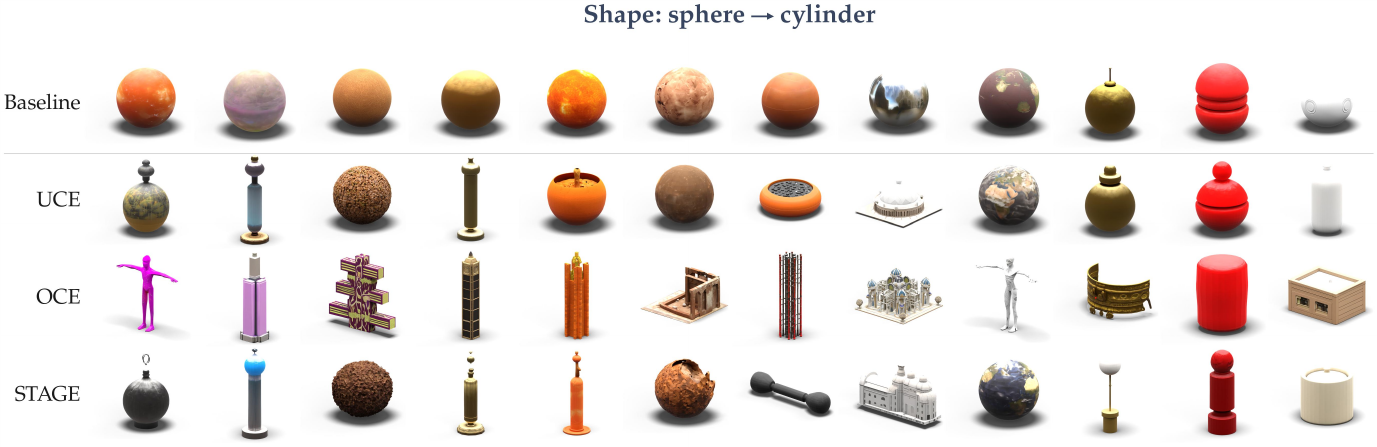}\\[0.5em]
      \includegraphics[width=0.95\textwidth]{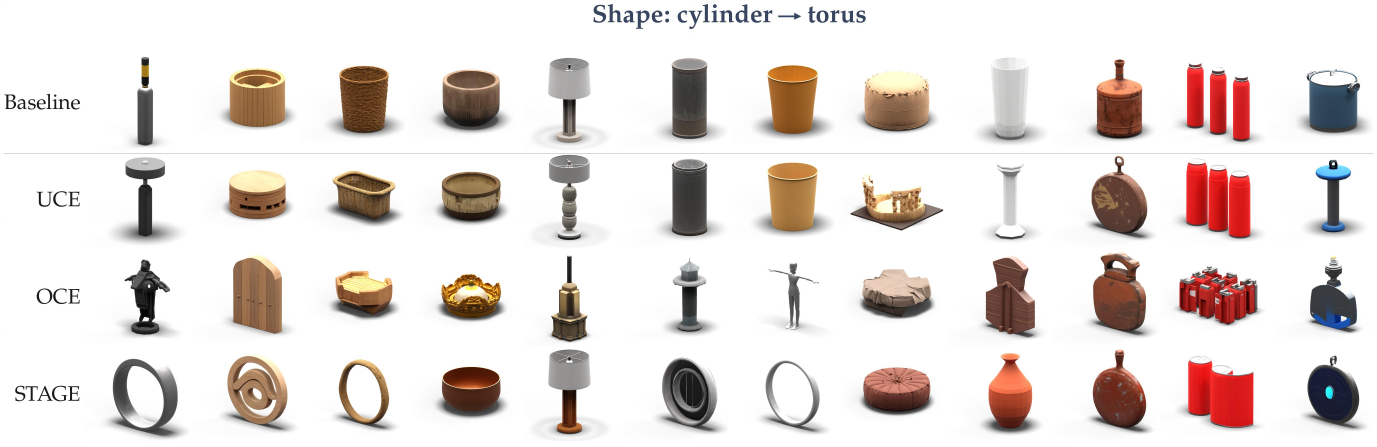}\\[0.5em]
      \includegraphics[width=0.95\textwidth]{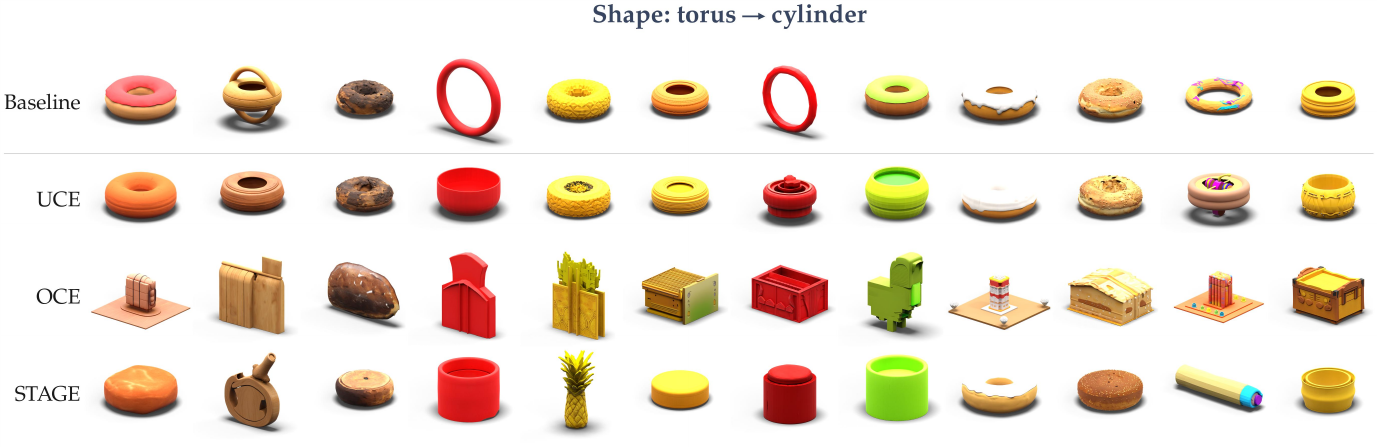}\\[0.5em]
      \includegraphics[width=0.95\textwidth]{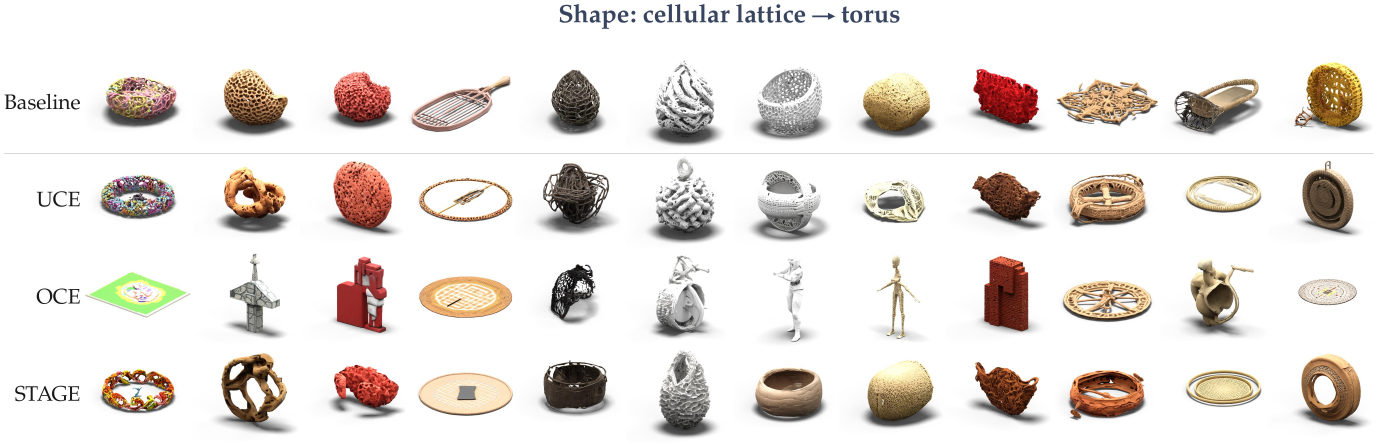}
      \caption{%
        Erasing shape concepts. The \emph{Baseline} row shows the unedited model, and the rows below show the same twelve prompts after erasure with UCE, OCE, and \our{}. Models share a scale within a column, not across columns.%
      }
      \label{fig:strips_shape}
    \end{figure}

Figures~\ref{fig:strips_material}, \ref{fig:strips_object}, and~\ref{fig:strips_shape} show twelve evaluation prompts, each with its fixed seed, for every concept of the material, object, and shape axes. For each concept, the first row shows the unedited model, and the other rows show the result of UCE, OCE, and \our{} with the per-axis stage assignment.

\end{document}

%% file: math_commands.tex
\usepackage{amsmath,amsfonts,bm}

\def\eqref#1{equation~\ref{#1}}

\def\1{\bm{1}}

\DeclareMathAlphabet{\mathsfit}{\encodingdefault}{\sfdefault}{m}{sl}
\SetMathAlphabet{\mathsfit}{bold}{\encodingdefault}{\sfdefault}{bx}{n}

